\documentclass[11pt]{article}
\PassOptionsToPackage{table}{xcolor}
\usepackage{booktabs}
\usepackage{multirow}
\usepackage{array}
\usepackage{amsmath}
\usepackage{amsfonts}
\usepackage{graphicx}
\usepackage{makecell}
\usepackage{subcaption}
\usepackage{graphicx}
\usepackage{float}
\usepackage[table]{xcolor}
\usepackage{capt-of}
\usepackage{placeins}
\usepackage{tabularx}
\usepackage{threeparttable}

\usepackage[final]{acl}

\usepackage{times}
\usepackage{latexsym}

\usepackage[T1]{fontenc}

\usepackage[utf8]{inputenc}

\usepackage{microtype}

\usepackage{inconsolata}

\usepackage{graphicx}
\usepackage{comment}

\title{EM\textsuperscript{2}Mem: Event-Centric Multimodal Memory for Large Language Models}

\author{
    Yijun Chen\textsuperscript{$1$}\thanks{Equal Contribution.}, 
    Yaqi Zheng\textsuperscript{$1$}\footnotemark[1], 
    Yanya Li\textsuperscript{$1$}, 
    Boyi Xiao\textsuperscript{$2$}, \\
    \textbf{Buqiang Xu\textsuperscript{$1$}},
    \textbf{Shuofei Qiao\textsuperscript{$1$}}, 
    \textbf{Jizhan Fang\textsuperscript{$1$}}, 
    \textbf{Xinle Deng\textsuperscript{$1$}}, 
    \textbf{Yunzhi Yao\textsuperscript{$1$}}, \\
    \textbf{Xuehai Wang\textsuperscript{$1$}},
    \textbf{Liuxin Zhang\textsuperscript{$3$}}, 
    \textbf{Hui Li\textsuperscript{$3$}}, 
    \textbf{Huajun Chen\textsuperscript{$1$}}, 
    \textbf{Shumin Deng\textsuperscript{$1$}\thanks{Corresponding Author.}} \\
    \textsuperscript{$1$}Zhejiang University \quad 
    \textsuperscript{$2$}South China University of Technology \\
    \textsuperscript{$3$}Lenovo Group Limited \\
    \texttt{ie\_yijunchen@zju.edu.cn, \quad 231sm@zju.edu.cn} \\
}

\begin{document}
\maketitle

\begin{abstract}
Multimodal memory offers a scalable interface for long-video question answering, but existing methods often retrieve captions, frames, transcripts, summaries, or graph facts as isolated fragments.
Although searchable, such fragments are not generation-ready: language models must reconstruct cross-modal and temporal alignments at inference time, when context is limited and attribution is difficult.
We propose EM\textsuperscript{2}Mem, an event-centric multimodal memory framework that binds heterogeneous evidence to event anchors during memory construction.
Each event-indexed memory cell aligns multimodal records, temporal context, graph-linked relations, semantic facts, and provenance, enabling compact evidence readout over grounded multimodal events rather than modality-specific fragments.
Across three long-video QA benchmarks, EM\textsuperscript{2}Mem improves average accuracy over the strongest memory baseline by 2.0, 2.4, and 3.7 points, improves strict event-level Top-5 evidence recall by 7.0 points, and reduces per-query latency by $4.67\times$ and total inference tokens by $63.66\%$\footnote{The code will be integrated into \url{https://github.com/zjunlp/LightMem}.
}.
\end{abstract}

\section{Introduction}
Multimodal memory is becoming essential for enabling Large Language Models (LLMs) to reason over extended multimodal experiences. 
Unlike short-image or short-video understanding, long-video QA requires models to answer questions whose supporting evidence may appear sparsely across minutes or hours of visual observations, speech transcripts, OCR, scenes, objects, and recurring behaviors \cite{EgoLife,Ego-R1,Video-MME}. 
Since such evidence cannot be reliably captured by a single context window or a flat video representation, LLMs need an external memory that can store observations, retrieve task-relevant context, and support grounded generation. 
The key problem, is not only how much information to store, but how to organize heterogeneous evidence into retrievable units that align with the way questions are asked and answers are justified.

\begin{figure}[!t]
    \centering
    \includegraphics[width=\columnwidth]{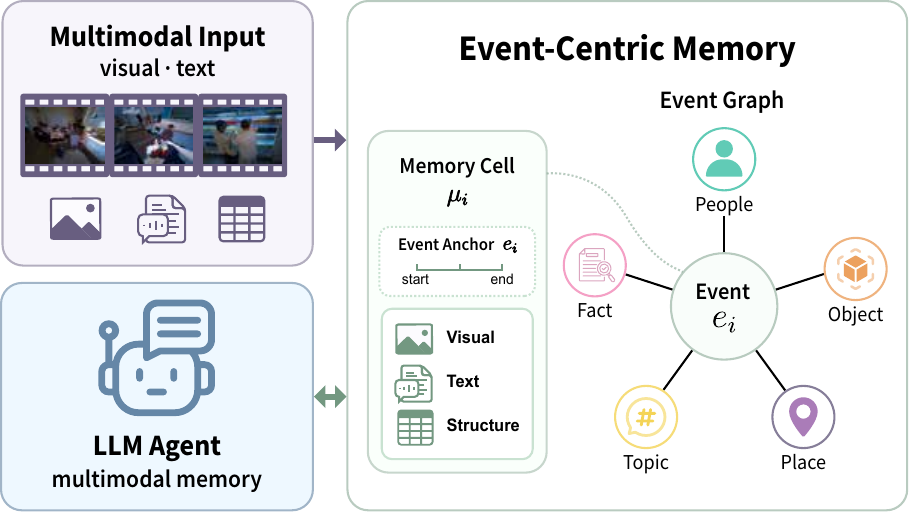}
    \caption{EM\textsuperscript{2}Mem organizes multimodal evidence into event-centric memory cells, enabling grounded retrieval over events rather than isolated fragments.}
    \label{fig:intro_comparison}
\end{figure}

Recent multimodal memory and video retrieval-augmented generation methods typically construct fragment-centric memories by compressing videos into modality- or structure-specific stores, such as captions, summaries, keyframes, transcript segments, embeddings, triples, or knowledge graphs \cite{Video-RAG__Visually-aligned,LangRepo,DrVideo,HippoRAG,WorldMM}. 
Although these fragments are searchable, they often fail to provide the event-level grounding needed for long-video QA. 
Questions commonly refer to events, situations, recurring behaviors, and temporally grounded relations, rather than isolated frames, sentences, or graph edges. 
As a result, existing systems face a \textit{retrieve-then-align} bottleneck: they first retrieve disconnected fragments and then rely on the LLM to reconstruct missing cross-modal and temporal bindings during generation, when the context budget is limited and attribution is the most difficult to maintain.

To address this bottleneck, we rethink the retrieval unit of multimodal memory: \textbf{\textit{How should multimodal memory organize heterogeneous evidence for LLM agents?}} 
Inspired by event cognition, which views events as coherent units for organizing perception, language, actions, and temporal boundaries \cite{zacks2007event,radvansky2014event,CLIP-Event}, we propose \textbf{EM\textsuperscript{2}Mem}, an \textbf{Event-Centric Multimodal Memory} framework for long-video QA. 
Rather than treating events as merely temporal video segments, EM\textsuperscript{2}Mem uses event anchors as language-addressable memory indices that bind captions, transcripts, keyframes, structured metadata, temporal context, and provenance during memory construction. 
This yields an \textit{align-then-retrieve} design: heterogeneous evidence is unified into event-indexed multimodal memory cells before retrieval, so the model retrieves grounded event-level evidence rather than assembling isolated fragments afterward. 
To support reasoning beyond individual events, EM\textsuperscript{2}Mem further builds an episodic graph for cross-event relations and temporal transitions, together with a semantic graph for long-term facts, habits, preferences, and stable relations, each grounded in supporting event evidence.

This event-centric design yields consistent gains in accuracy, efficiency, and attribution. 
On three long-video QA benchmarks, EgoLifeQA, Ego-R1 Bench, and Video-MME (L), EM\textsuperscript{2}Mem improves average accuracy over the strongest memory baseline by 2.0\%, 2.4\%, and 3.7\% respectively, while reducing per-query latency by 4.67 times and total inference tokens by 63.66\%. 
It also localizes evidence more precisely, improving strict event-level Top-5 recall by 7.0 points over WorldMM's five rounds of iterative retrieval. 
Further analyses show that structured event-centric evidence provides a more effective memory interface than raw frames or flattened captions, and that construction-time unification outperforms retrieval-time fusion. 
Together, these results suggest that organizing multimodal memory around events provides a more natural and effective foundation for grounded, attributable long-video QA.

\section{Background}

Given a long video with visual and text-form evidence streams \(V=\{V^{\mathrm{vis}},V^{\mathrm{text}}\}\) and a question \(q\), long-video QA aims to generate an answer \(\hat{y}\) grounded in video-specific evidence.
Here, \(V^{\mathrm{text}}\) includes transcripts, OCR, subtitles, dialogue, and other textual cues \cite{Video-RAG__Visually-aligned, VideoLLaMA_2, VAST}.
Since relevant evidence is often sparse, temporally distant, and distributed across modalities, directly feeding the full video into an LLM or MLLM is inefficient and may obscure cross-modal dependencies.

Existing multimodal memory systems typically convert long videos or agent experiences into searchable modality-typed units \cite{WorldMM, M3-Agent, VideoAgent, Omni-SimpleMem}:
\begin{equation}
    \begin{aligned}
        u_j^{m} &= \left(s_j^{m}, e_j^{m}, p_j^{m}, \tau_j, m, \ell_j\right), \quad m\in\{\mathrm{vis},\mathrm{text}\}, \\
        s_j^{\mathrm{vis}} &= c_j^{\mathrm{clip}}=\phi_{\mathrm{cap}}(F_j,K_j), \\
        s_j^{\mathrm{text}} &= \phi_{\mathrm{text}}(r_j,o_j,d_j).
    \end{aligned}
\end{equation}
Here, \(s_j^{m}\) and \(e_j^{m}\) denote the searchable summary and embedding, \(p_j^{m}\) points to the source evidence, \(\tau_j\) is the timestamp, and \(\ell_j\) stores links to other units.
Visual frames \(F_j\) and keyframes \(K_j\) are converted into clip captions, while transcripts \(r_j\), OCR/subtitles \(o_j\), and other textual cues \(d_j\) are normalized into text summaries.
Although such units provide a unified storage interface, they remain organized around individual modality-specific signals or compressed observations rather than coherent multimodal events.

\section{Method}

\begin{figure*}[!t]
    \centering
    \includegraphics[width=\textwidth]{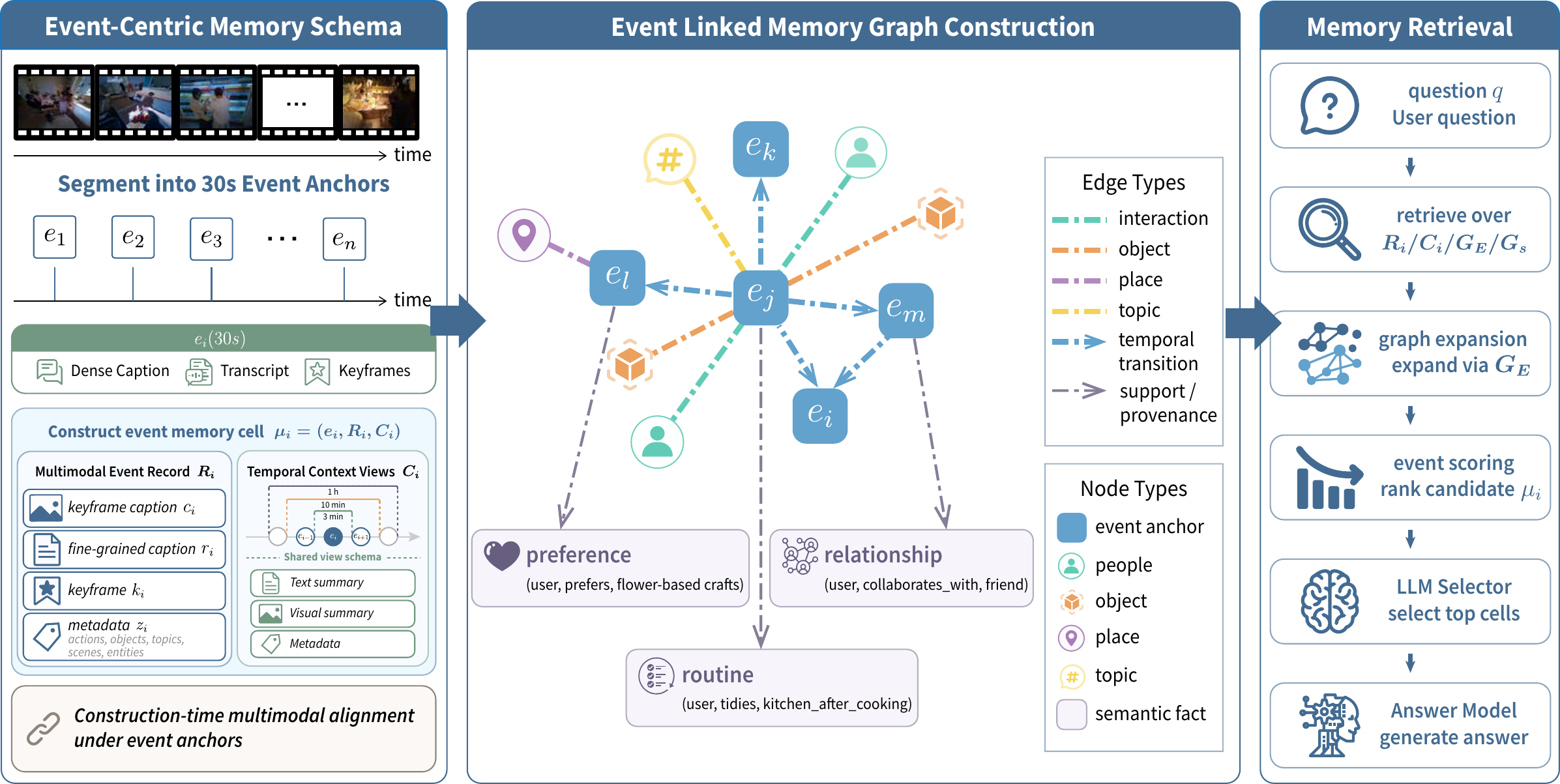}
    \caption{
    Overview of EM\textsuperscript{2}Mem.
    EM\textsuperscript{2}Mem parses a long video into event anchors, constructs event-centric multimodal memory cells with local records \(R_i\) and multi-scale context views \(\mathcal{C}_i\), and links them through episodic and semantic graphs \(G_E\) and \(G_S\).
    Given a question, lightweight retrieval selects and expands relevant event cells to compile evidence for answer generation.
    }
    \label{fig:framework}
    \vspace{-3mm}
\end{figure*}


\subsection{Overview}
We propose \textbf{EM\textsuperscript{2}Mem}, an \textbf{Event-Centric Multimodal Memory} framework for long-video reasoning with LLMs. EM\textsuperscript{2}Mem follows an \textit{align-then-retrieve} design for long-video memory, as illustrated in Figure~\ref{fig:framework}. 
It first parses a long video into short temporal events and uses event anchors as shared indices to align heterogeneous evidence during memory construction, rather than retrieving isolated modality-specific fragments at inference time. 
To support reasoning beyond individual events, EM\textsuperscript{2}Mem further builds lightweight event-linked graph memory: an episodic graph captures concrete cross-event relations, including shared entities, objects, scenes, topics, and temporal transitions, while a semantic graph stores higher-level regularities grounded in supporting events. 
At inference time, EM\textsuperscript{2}Mem retrieves relevant event cells, expands them with graph-linked evidence, and compiles a compact query-specific evidence view for answer generation.

\subsection{Event-Centric Multimodal Memory Schema}
Instead of treating a video as a flat sequence of frames or captions, EM\textsuperscript{2}Mem represents it as an event-indexed memory.
We divide the video into short base segments, e.g. 30-second clips, and associate each segment with an event anchor:
\begin{equation}
    e_i=\left(i,\tau_i^s,\tau_i^e\right).
\end{equation}

An event anchor is a temporal address rather than memory content.
It provides a shared indexing key for heterogeneous evidence and higher-level abstractions.
Formally, EM\textsuperscript{2}Mem represents the video as a set of \textbf{Event-Centric Multimodal Memory Cells}:
\begin{equation}
    \begin{aligned}
         \mathcal{M}_{\mathrm{EM}^{2}\mathrm{Mem}}
        &=
        \left(
        \{\mu_i\}_{i=1}^{N},
        G_E,
        G_S
        \right), \\
        \mu_i&=(e_i,R_i,\mathcal{C}_i).   
    \end{aligned}
\end{equation}
Each event memory cell \(\mu_i\) is anchored at \(e_i\) and stores local multimodal evidence \(R_i\) and multi-scale temporal context views \(\mathcal{C}_i\).
Beyond cell-level memory, EM\textsuperscript{2}Mem builds two global event-linked graphs: \(G_E\) captures cross-event episodic relations and temporal transitions, while \(G_S\) captures long-term semantic relations.

To make this schema concrete, suppose an event segment shows a person discussing a flower-based craft plan while interacting with flowers and vases.
The corresponding memory cell binds the dialogue context, keyframe captions, and structured fields such as \(\mathrm{Act}_i=\textit{discussing a craft plan}\), \(\mathrm{Obj}_i=\{\textit{flowers},\textit{vases}\}\), and \(\mathrm{Top}_i=\textit{flower-based craft}\) under the same event anchor \(e_i\), rather than storing them as separate captions, transcript snippets, and frames.
Its context views \(\mathcal{C}_i\) link the event to surrounding temporal summaries, while the graphs connect it to related entities, objects, topics, and recurring patterns.
This design makes the memory cell query-ready: questions about the plan, involved objects, or nearby activities can retrieve one grounded event cell and its linked context instead of stitching together isolated modality-specific fragments at inference time.
All evidence is grounded to event anchors via \(\alpha(\cdot)\), preserving heterogeneous memory forms within a event-centric schema.

\subsection{Multimodal Memory Construction}
\paragraph{Multimodal Event Record.}
For each base event segment, EM\textsuperscript{2}Mem constructs a multimodal event record that stores the local evidence observed within the event:
\begin{equation}
    \label{eq:region-representation}
    \begin{aligned}
        R_i &= \bigl(c_i, r_i, k_i, z_i, \tau_i\bigr), \\
        z_i &= \bigl(\mathrm{Act}_i, \mathrm{Obj}_i, \mathrm{Top}_i,
                \mathrm{Scen}_i, \mathrm{Ent}_i\bigr).
    \end{aligned}
\end{equation}
Here, \(c_i\) is a visual keyframe caption, \(r_i\) is the transcript or dialogue context, \(k_i\) denotes representative keyframes, and \(z_i\) contains structured metadata, including actions, objects, dialogue topics, visual scenes, and visual entities. 
These fields are obtained through multimodal parsing modules, including captioning, transcript alignment, keyframe selection, and structured field extraction.
\paragraph{Temporal Context Views.}
To support reasoning at different temporal granularities, EM\textsuperscript{2}Mem further builds temporal context views over multiple scales, such as 3 minutes, 10 minutes, and 1 hour. 
At each scale, consecutive event anchors are grouped into a temporal block and summarized into a context view:
\begin{equation}
    h_j^{(l)}
    =(u_{j,l}^{\mathrm{text}},u_{j,l}^{\mathrm{vis}},m_{j,l},\tau_{j,l},l).
\end{equation}
The textual summary \(u_{j,l}^{\mathrm{text}}\) captures narrative and dialogue context, the visual summary \(u_{j,l}^{\mathrm{vis}}\) summarizes visual evidence, and \(m_{j,l}\) aggregates normalized metadata such as actions, objects, scenes, and topics. 
Each context view is anchored to all base events covered by its temporal span, and the context-view set \(\mathcal{C}_i\) of an event cell contains all views whose spans include \(e_i\).

\subsection{Event-Linked Memory Graph Construction}

Beyond event memory cells, EM\textsuperscript{2}Mem builds two lightweight graph indices over shared event anchors: an episodic graph \(G_E\) and a semantic graph \(G_S\).
These graphs are used as auxiliary indices rather than as new graph formalisms. 
The episodic graph connects event anchors with typed entities such as people, objects, locations, scenes, and topics, and also includes temporal transitions between adjacent events. Each relation is grounded to its supporting event anchors, allowing graph-derived clues to be traced back to concrete video evidence.

The semantic graph summarizes recurring patterns from episodic evidence, such as habits, preferences, routines, and stable relationships. 
Unlike local event records, which describe concrete observations, semantic facts capture long-term regularities that may span multiple distant events. 
These facts are also linked back to their supporting anchors, so that high-level knowledge remains evidence-grounded. Together, \(G_E\) and \(G_S\) complement event memory cells by providing cross-event connectivity and long-term semantic
context.

\subsection{Memory Retrieval and Ranking}

At inference time, EM\textsuperscript{2}Mem performs lightweight event-level readout over pre-aligned memory cells, which shifts retrieval from query-time cross-modal fusion to compact event-cell selection and expansion.
Given a question \(q\), it first identifies relevant event memory cells using signals from local multimodal records, multi-scale temporal context views, and graph-linked evidence.
It then performs lightweight expansion through temporal transitions, shared entities, objects, locations, topics, and relevant semantic facts, so that complementary evidence from nearby or semantically related events can be included.

An LLM-based selector further filters the candidate event cells and keeps only the evidence needed to answer the question.
For each selected event, EM\textsuperscript{2}Mem compiles a query-specific evidence view \(E_q\), including relevant captions, transcripts, structured visual fields, temporal summaries, semantic facts, and selected keyframes.
The final answer \(\hat{y}\) is generated by \(\operatorname{LLM}_{\mathrm{ans}}\) conditioned on the question \(q\) and the compact evidence view \(E_q\).

\section{Experiment}
\subsection{Experimental Setup}
\paragraph{Datasets and Metrics.}
We evaluate EM\textsuperscript{2}Mem on three multiple-choice long-video QA benchmarks: EgoLifeQA, Ego-R1 Bench, and Video-MME (L) \cite{EgoLife, Ego-R1, Video-MME}.
Together, they cover week-long egocentric daily-life QA, ultra-long egocentric reasoning, and open-domain long-video understanding. Following the standard protocol of each benchmark, we report category-level and average accuracy. Dataset statistics, category definitions, and license details are provided in Appendix \ref{app:dataset}.

\paragraph{Baselines.}
We compare EM\textsuperscript{2}Mem with representative baselines spanning base MLLMs, long-video MLLMs, RAG-based video QA methods, and memory-based long-video reasoning frameworks, as listed in Tables~\ref{tab:egolifeqa_egor1bench} and~\ref{tab:video_mme_l}. 
For the strongest and most related memory baseline, we report both the originally published WorldMM results and our reproduced WorldMM$^\dagger$ results under the same evaluation setting as EM\textsuperscript{2}Mem. 
Detailed baseline configurations are provided in Appendix~\ref{app:baseline-setup}.


\subsection{Main Results}
\label{sec:main_results}
\begin{table*}[t]
\centering
\small
\setlength{\tabcolsep}{4.5pt}
\renewcommand{\arraystretch}{1.08}

\begin{tabular}{
l
ccccc>{\columncolor{gray!15}}c
ccccc>{\columncolor{gray!15}}c
}
\toprule[1.2pt]
\multirow{2}{*}{\textbf{Model}} 
& \multicolumn{6}{c}{\textbf{EgoLifeQA}} 
& \multicolumn{6}{c}{\textbf{Ego-R1 Bench}} \\
\cmidrule(lr){2-7}\cmidrule(lr){8-13}
& \textbf{Ent.} & \textbf{EvR.} & \textbf{Hab.} & \textbf{Rel.} & \textbf{Task} & \textbf{Avg.}
& \textbf{Ent.} & \textbf{EvR.} & \textbf{Hab.} & \textbf{Rel.} & \textbf{Task} & \textbf{Avg.} \\
\midrule

\multicolumn{13}{l}{\textbf{\textit{Base Models}}} \\
Qwen3-VL-8B    & 35.2 & 30.2 & 39.3 & 46.4 & 46.0 & 38.6 & 31.8 & 41.5 & 38.5 & 42.1 & 44.7 & 35.7 \\
Gemini 2.5 Pro & 43.2 & 40.5 & 41.0 & 55.2 & 52.4 & 46.4 & 43.9 & 56.1 & 53.9 & 47.4 & 47.4 & 46.7 \\
GPT-5          & 47.2 & 42.1 & 47.5 & 53.6 & 55.6 & 48.6 & 41.8 & 58.5 & 53.9 & 52.6 & 50.0 & 46.3 \\
\midrule

\multicolumn{13}{l}{\textbf{\textit{Long Video LLMs}}} \\
VideoChat-Flash & 28.8 & 32.5 & 37.7 & 37.6 & 38.1 & 34.2 & 43.4 & 43.9 & 38.5 & 31.6 & 44.7 & 42.7 \\
Time-R1         & 39.2 & 50.8 & 65.6 & 48.8 & 47.6 & 48.8 & 49.2 & 48.8 & 46.2 & 42.1 & 44.7 & 48.0 \\
Video-RTS       & 40.8 & 48.4 & 62.3 & 48.8 & 47.6 & 48.2 & 47.6 & 46.3 & 53.9 & 52.6 & 47.4 & 48.0 \\
\midrule

\multicolumn{13}{l}{\textbf{\textit{RAG-based Video LLMs}}} \\
LightRAG   & 40.8 & 48.4 & 67.2 & 50.4 & 44.4 & 48.8 & 54.0 & 61.0 & 46.2 & 42.1 & 42.1 & 52.3 \\
HippoRAG   & 48.8 & 60.3 & 70.5 & 60.8 & 66.7 & 59.6 & 54.5 & 65.9 & 69.2 & 52.6 & 50.0 & 56.0 \\
Video-RAG  & 49.6 & 56.3 & 67.2 & 55.2 & 54.0 & 55.4 & 48.7 & 58.5 & 53.9 & 47.4 & 44.7 & 49.7 \\
\midrule

\multicolumn{13}{l}{\textbf{\textit{Memory-based Video LLMs}}} \\
EgoRAG        & 40.0 & 56.3 & 62.3 & 54.4 & 52.4 & 52.0 & 46.6 & 56.1 & 46.2 & 47.4 & 55.3 & 49.0 \\
Ego-R1        & 51.2 & 53.2 & 63.9 & 50.4 & 50.8 & 53.0 & 50.8 & 63.4 & 38.5 & 36.8 & 57.9 & 52.0 \\
HippoMM       & 45.6 & 53.2 & 70.5 & 55.2 & 58.7 & 54.6 & 51.9 & 56.1 & 46.2 & 52.6 & 57.9 & 53.0 \\
M3-Agent      & 44.4 & 54.8 & 62.3 & 56.8 & 54.0 & 53.5 & 52.4 & 58.5 & 38.5 & 42.1 & 52.6 & 52.0 \\
WorldMM       & \textbf{62.4} & 64.3 & \textbf{75.4} & 62.4 & 71.4 & 65.6 & 64.6 & \textbf{70.7} & \textbf{76.9} & \textbf{57.9} & \textbf{63.2} & 65.3 \\
WorldMM$^\dagger$ 
              & 57.6 & \textbf{65.1} & 68.9 & 68.8 & 60.3 & 64.0 
              & \multicolumn{6}{c}{--} \\
\midrule

\rowcolor{blue!8}
\rowcolor{blue!8}
\textbf{EM\textsuperscript{2}Mem} \textbf{\textit{(Ours)}} 
              & 60.8 & 61.1 & 63.9 & \textbf{72.8} & \textbf{74.6} & \textbf{66.0} 
              & \textbf{74.6} & 53.7 & 69.2 & 47.4 & 57.9 & \textbf{67.7} \\
\bottomrule[1.2pt]
\end{tabular}%
\vspace{-1mm}
\caption{
Performance comparison on EgoLifeQA and Ego-R1 Bench.
WorldMM denotes the originally reported results.
$^\dagger$ denotes our reproduced results under the same evaluation setting as EM\textsuperscript{2}Mem.
For Ego-R1 Bench, we report the published WorldMM results because reproducing WorldMM requires rebuilding subject-level long-video memories for all six participants, which incurs substantial construction and inference cost. The corresponding entries are marked as ``--''.
}
\label{tab:egolifeqa_egor1bench}
\vspace{-3mm}
\end{table*}

\begin{table}[!htbp]
\centering
\footnotesize
\setlength{\tabcolsep}{3.8pt}
\begin{tabular*}{\columnwidth}{@{\extracolsep{\fill}}lrrr@{}}
\toprule
\textbf{Metric} & \textbf{WorldMM} & \textbf{EM\textsuperscript{2}Mem} & \textbf{Gain} \\
\midrule
\multicolumn{4}{@{}l}{\textit{Per-query inference efficiency}} \\
Avg. latency / query (s) & 459.00 & \textbf{98.21} & \textbf{4.67$\times$} \\
\midrule
\multicolumn{4}{@{}l}{\textit{End-to-end evaluation throughput}} \\
Wall-clock time (s) & 229,502 & \textbf{6,138} & \textbf{37.39$\times$} \\
\midrule
\multicolumn{4}{@{}l}{\textit{Inference token cost}} \\
Input tokens  & 31.95M & \textbf{13.29M} & \textbf{58.42\%$\downarrow$} \\
Output tokens & 10.08M & \textbf{1.99M}  & \textbf{80.29\%$\downarrow$} \\
Total tokens  & 42.03M & \textbf{15.27M} & \textbf{63.66\%$\downarrow$} \\
\bottomrule
\end{tabular*}
\vspace{-1mm}
\caption{
Inference efficiency comparison on EgoLifeQA.
Latency is measured per query; wall-clock time follows each method's practical evaluation pipeline.
}
\label{tab:efficiency}
\vspace{-3mm}
\end{table}

\begin{table*}[t]
\centering
\small
\setlength{\tabcolsep}{4.2pt}
\renewcommand{\arraystretch}{1.08}

\resizebox{\textwidth}{!}{%
\begin{tabular}{
l
cccccccccccc>{\columncolor{gray!15}}c
}
\toprule[1.2pt]
\textbf{Model} 
& \textbf{ARES} & \textbf{AREC} & \textbf{ATTR} & \textbf{CNT} & \textbf{ISYN} & \textbf{OCR}
& \textbf{ORES} & \textbf{OREC} & \textbf{SPER} & \textbf{SRES} & \textbf{TPER} & \textbf{TRES}
& \textbf{Avg.} \\
\midrule

\multicolumn{14}{l}{\textbf{\textit{Base Models}}} \\
Qwen3-VL-8B    & 62.2 & 54.0 & 51.9 & 43.8 & 68.1 & 42.9 & 62.9 & 57.4 & 33.3 & 45.5 & 33.3 & 67.0 & 61.0 \\
Gemini 2.5 Pro & 56.9 & 47.6 & 66.7 & 41.7 & 71.8 & 57.1 & 53.3 & 40.7 & 0.0  & 72.7 & \textbf{66.7} & 48.4 & 55.7 \\
GPT-5          & 71.1 & 69.8 & 70.4 & 47.9 & \textbf{88.3} & 57.1 & 75.8 & 74.1 & 33.3 & 72.7 & 50.0 & 75.8 & 74.3 \\
\midrule

\multicolumn{14}{l}{\textbf{\textit{Long Video LLMs}}} \\
VideoChat-Flash & 35.0 & 42.9 & 37.0 & 31.3 & 34.4 & 42.9 & 60.0 & 46.3 & 33.3 & 54.5 & 33.3 & 46.2 & 44.1 \\
Time-R1         & 20.6 & 28.6 & 25.9 & 35.4 & 31.9 & 35.7 & 53.3 & 48.2 & 33.3 & 36.4 & 50.0 & 44.0 & 37.6 \\
Video-RTS       & 43.3 & 52.4 & 40.7 & 39.6 & 33.7 & 42.9 & 60.8 & 53.7 & 33.3 & 45.5 & 50.0 & 49.5 & 47.9 \\
\midrule

\multicolumn{14}{l}{\textbf{\textit{RAG-based Video LLMs}}} \\
LightRAG  & 41.7 & 30.2 & 40.7 & 35.4 & 54.0 & 50.0 & 46.7 & 61.1 & 33.3 & 45.5 & 50.0 & 52.8 & 46.6 \\
HippoRAG  & 45.6 & 47.6 & 40.7 & 37.5 & 52.2 & 42.9 & 52.9 & 64.8 & 66.7 & 54.5 & 50.0 & 70.3 & 52.1 \\
Video-RAG & 51.7 & 47.6 & 37.0 & 39.6 & 49.7 & 57.1 & 62.1 & 68.5 & 66.7 & 45.5 & 50.0 & 68.1 & 55.4 \\
\midrule

\multicolumn{14}{l}{\textbf{\textit{Memory-based Video LLMs}}} \\
EgoRAG   & 31.1 & 55.6 & 33.3 & 22.9 & 41.1 & 28.6 & 44.6 & 48.2 & 33.3 & 54.5 & \textbf{66.7} & 48.4 & 41.1 \\
Ego-R1   & 37.2 & 52.4 & 40.7 & 35.4 & 38.0 & 35.7 & 42.1 & 51.9 & \textbf{66.7} & 63.6 & 50.0 & 52.8 & 42.7 \\
HippoMM  & 41.1 & 42.9 & 55.6 & 35.4 & 38.7 & 35.7 & 37.9 & 53.7 & 33.3 & 54.5 & 50.0 & 47.3 & 41.6 \\
M3-Agent & 52.2 & 57.1 & 59.3 & 45.8 & 51.5 & 42.9 & 54.6 & 64.8 & 33.3 & 45.5 & 50.0 & 71.4 & 55.3 \\
WorldMM  & \textbf{81.1} & 73.0 & 70.4 & 54.2 & 85.3 & 42.9 & 75.0 & \textbf{77.8} & 33.3 & 72.7 & \textbf{66.7} & \textbf{79.1} & 76.6 \\
WorldMM$^\dagger$ 
          & 73.3 & 68.3 & \textbf{77.8} & 60.4 & 80.2 & 50.0 & 72.4 & \textbf{77.8} & 33.3 & \textbf{90.9} & \textbf{66.7} & 71.1 & 73.1 \\
\midrule

\rowcolor{blue!8}
\rowcolor{blue!8}
\textbf{EM\textsuperscript{2}Mem} \textbf{\textit{(Ours)}} 
          & 77.2 & \textbf{76.2} & \textbf{77.8} & \textbf{64.6} & 80.7 & \textbf{64.3} 
          & \textbf{77.0} & \textbf{77.8} & 33.3 & 81.8 & 50.0 & \textbf{79.1} & \textbf{76.8} \\
\bottomrule[1.2pt]
\end{tabular}%
}
\vspace{-1mm}
\caption{
Category-wise performance on Video-MME (L).
WorldMM denotes the originally reported results.
$^\dagger$ denotes our reproduced results under the same evaluation setting as EM\textsuperscript{2}Mem.
}
\label{tab:video_mme_l}
\end{table*}
\paragraph{Accuracy across long-video QA benchmarks.}
Tables~\ref{tab:egolifeqa_egor1bench} and~\ref{tab:video_mme_l} report the main results on EgoLifeQA, Ego-R1 Bench, and Video-MME (L).
Overall, EM\textsuperscript{2}Mem achieves strong performance against base MLLMs, long-video LLMs, RAG-based methods, and memory-based video reasoning systems.
We report both previously published WorldMM results and our reproduced WorldMM$^\dagger$ results for reference; since reproduced results are obtained under the same evaluation setting as EM\textsuperscript{2}Mem, they provide the most controlled comparison.
Under this setting, EM\textsuperscript{2}Mem improves over WorldMM$^\dagger$ by 2.0 points on EgoLifeQA (66.0 vs. 64.0), and 3.7 points on Video-MME (L) (76.8 vs. 73.1).
On Ego-R1 Bench, where we use the originally reported WorldMM result for comparison, EM\textsuperscript{2}Mem achieves 67.7 average accuracy, outperforming WorldMM by 2.4 points (67.7 vs. 65.3).
These gains are especially visible on RelationMap and TaskMaster in EgoLifeQA, EntityLog in Ego-R1 Bench, and AREC, CNT, OCR, and ORES in Video-MME (L), suggesting that event-indexed retrieval units are useful for grounded entity tracking, task-state reasoning, and cross-modal evidence aggregation.
Compared with the originally reported WorldMM numbers, EM\textsuperscript{2}Mem remains competitive and slightly higher on average, although WorldMM is stronger on several habit, temporal, and synthetic reasoning categories.

\paragraph{Inference efficiency and scalability.}
Beyond accuracy, Table~\ref{tab:efficiency} compares EM\textsuperscript{2}Mem with WorldMM along three efficiency dimensions: per-query latency, end-to-end wall-clock time, and inference token cost. 
We report both latency and wall-clock time because they capture different aspects of efficiency: latency measures the delay of an individual request, while wall-clock time reflects end-to-end throughput under each method's practical execution pipeline. 
In this setting, EM\textsuperscript{2}Mem processes independent questions with 8 workers, whereas WorldMM is constrained by incremental HippoRAG graph construction during inference.
EM\textsuperscript{2}Mem reduces average latency from 459.00s to 98.21s, yielding a \(4.67\times\) per-query speedup. 
This shows that the efficiency gain is not only due to parallel execution. 
The main reason is that EM\textsuperscript{2}Mem reads from pre-constructed event-indexed memory cells instead of performing evidence alignment and graph organization during inference.
EM\textsuperscript{2}Mem further reduces wall-clock evaluation time from 229,502s to 6,138s, corresponding to a \(37.39\times\) throughput improvement, and lowers total inference token cost from 42.03M to 15.27M, a 63.66\% reduction. 
These results indicate that construction-time memory organization improves not only answer quality, but also the inference-time cost of evidence readout.
Figure~\ref{fig:analysis_amortization} further contextualizes these savings by amortizing the offline construction cost over repeated queries on the same memory.
Although EM\textsuperscript{2}Mem incurs a higher upfront construction cost, its lower inference-time wall-clock cost reaches break-even after roughly 23--24 queries, while token usage requires a larger reuse scale to break even; we provide the full end-to-end derivation in Appendix~\ref{app:e2e_cost}.
Taken together, these results suggest that event-indexed retrieval units improve long-video QA accuracy while making inference-time evidence readout more scalable.

\subsection{Ablation Study}
Table ~\ref{tab:ablation_overall} and Figure~\ref{fig:analysis_component_drop} show the overall ablation results on EgoLifeQA.
Removing temporal context views leads to the largest drop, decreasing accuracy from 66.0\% to 60.4\%.
Removing semantic facts and episodic graph support also substantially hurts performance, reducing accuracy to 61.4\% and 61.6\%, respectively.
These results validate the design of combining local event records, temporal context views, episodic graph, and semantic facts in an event-indexed memory.

\begin{table}[!t]
\centering
\small
\begin{tabular}{lcc}
\toprule
\textbf{Method} & \textbf{Overall Acc.} & $\Delta$ \\
\midrule
EM${^2}$Mem & \textbf{66.0} & -- \\
w/o Temporal Context Views & 60.4 & $-5.6$ \\
w/o Semantic Memory & 61.4 & $-4.6$ \\
w/o Episodic Graph & 61.6 & $-4.4$ \\
Local 30s Event Record & 64.0 & $-2.0$ \\
\bottomrule
\end{tabular}
\vspace{-1mm}
\caption{
Overall ablation results on EgoLifeQA. 
$\Delta$ denotes the absolute accuracy drop compared with the full EM\textsuperscript{2}Mem model.
}
\label{tab:ablation_overall}
\end{table}

Interestingly, the local 30-second record retrieval baseline remains competitive, achieving 64.0\%.
This suggests that fine-grained local event record already provide strong evidence for short-range factual questions.
However, the full EM${^2}$Mem still achieves the best overall accuracy.
The category-wise results in Table \ref{tab:category_ablation} at Appendix \ref{app:ablation} show that the advantage of EM${^2}$Mem is more pronounced on relation- and task-oriented questions, where evidence often needs to be aggregated across temporally distant events and connected through entity, object, or activity relations.

\subsection{Analysis}
\label{sec:analysis}

\paragraph{Event binding turns heterogeneous evidence into answerable memory units.}
We analyze how the representation and fusion stage of multimodal evidence affect memory-based long-video QA, using the first 250 questions of EgoLifeQA.
Our analysis spans two dimensions: (1) memory representation of visual evidence, including raw frames, caption-level descriptions, and structured event fields; and (2) unification stage, comparing retrieval-time fusion, where independent stores are combined during inference, with construction-time unification, where cross-modal evidence is aligned under shared event anchors before indexing.

\begin{figure*}[t]
    \centering

    \begin{subfigure}[t]{0.318\textwidth}
        \centering
        \includegraphics[
            width=\linewidth,
            trim=3 3 3 3,
            clip
        ]{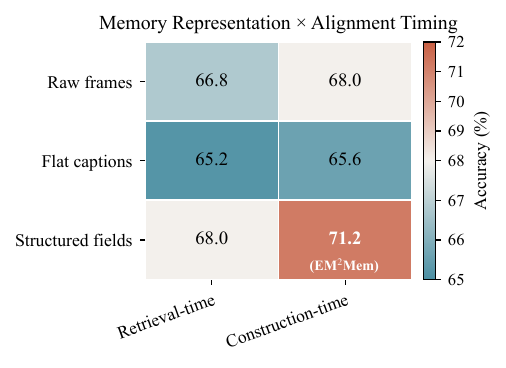}
        \caption{Representation and timing.}
        \label{fig:analysis_memory_alignment}
    \end{subfigure}
    \hfill
    \begin{subfigure}[t]{0.318\textwidth}
        \centering
        \includegraphics[
            width=\linewidth,
            trim=3 3 3 3,
            clip
        ]{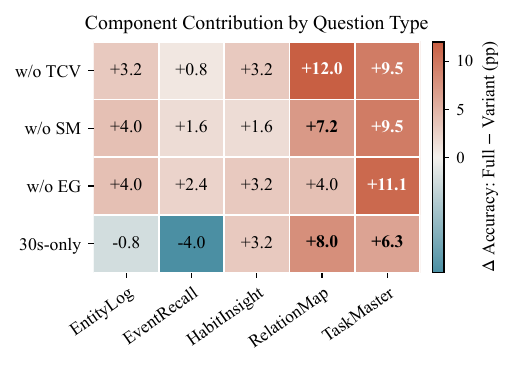}
        \caption{Component contribution.}
        \label{fig:analysis_component_drop}
    \end{subfigure}
    \hfill
    \begin{subfigure}[t]{0.318\textwidth}
        \centering
        \includegraphics[
            width=\linewidth,
            trim=3 3 3 3,
            clip
        ]{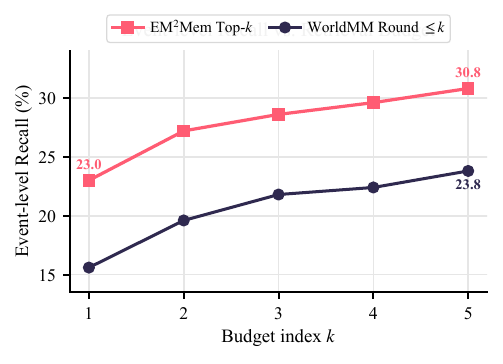}
        \caption{Recall vs. retrieval budget.}
        \label{fig:analysis_selector_budget}
    \end{subfigure}

    \vspace{1.5mm}

    \begin{subfigure}[t]{0.478\textwidth}
        \centering
        \includegraphics[
            width=\linewidth,
            trim=2 2 2 2,
            clip
        ]{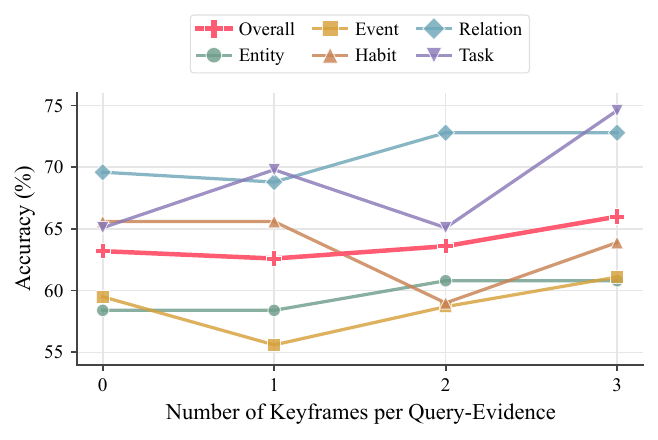}
        \caption{Accuracy vs. keyframe budget.}
        \label{fig:analysis_keyframes}
    \end{subfigure}
    \hfill
    \begin{subfigure}[t]{0.478\textwidth}
        \centering
        \includegraphics[
            width=\linewidth,
            trim=2 2 2 2,
            clip
        ]{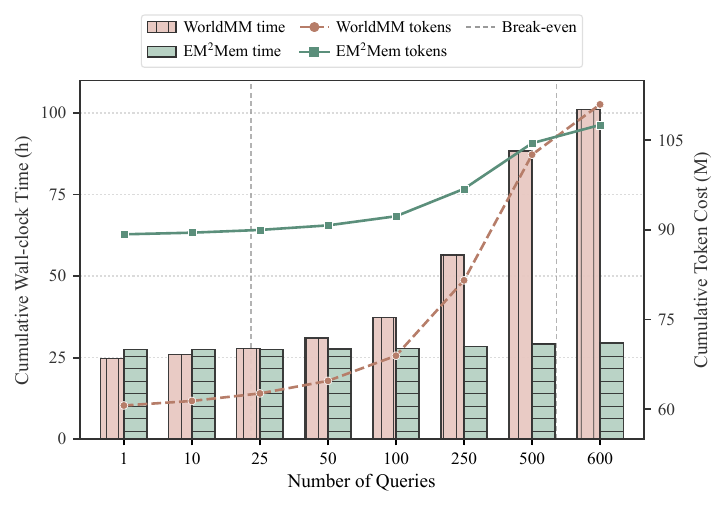}
        \caption{Amortized wall-clock and token cost.}
        \label{fig:analysis_amortization}
    \end{subfigure}


    \caption{
    Analysis of EM\textsuperscript{2}Mem design choices and efficiency.
    \textbf{(a)} Structured event fields with construction-time unification achieve the best accuracy.
    \textbf{(b)} Component-wise ablations show different contributions across question types.
    \textbf{(c)} EM\textsuperscript{2}Mem improves event-level recall as retrieval budget increases.
    \textbf{(d)} Visual keyframes improve overall accuracy, especially with three keyframes.
    \textbf{(e)} EM\textsuperscript{2}Mem amortizes its wall-clock overhead as the number of queries increases; dashed lines indicate wall-clock and token break-even points.
    }
    \label{fig:main_analysis}
\end{figure*}

\paragraph{Structured event fields bridge multimodal evidence and natural-language questions.} 
As shown in Table~\ref{tab:memory-unification} and Figure~\ref{fig:analysis_memory_alignment}, the representation of visual evidence has a substantial impact on QA performance. 
Across both fusion stages, structured evidence outperforms raw and caption-level memories, achieving 71.2\% accuracy when combined with construction-time unification. 
This suggests typed visual fields form a more effective memory interface for LLM-based QA: compared with raw frames, they are easier to retrieve, verbalize, and attribute; compared with flattened captions, they preserve explicit entities, actions, scenes, and other question-relevant cues. 
Rather than serving only as a visual representation, structured evidence acts as an answerable interface between multimodal observations and natural-language questions.

\vspace{2mm}
\begin{table}[!htbp]
\centering
\small
\setlength{\tabcolsep}{5pt}
\begin{tabular}{lcc}
\toprule
\multirow{2}{*}{\textbf{Memory Representation}} 
& \multicolumn{2}{c}{\textbf{Unification Stage}} \\
\cmidrule(lr){2-3}
& \textbf{Retrieval} & \textbf{Construction} \\
\midrule
Raw frames          & 66.8 & 68.0 \\
Flattened captions   & 65.2 & 65.6 \\
Structured event fields & 68.0 & \textbf{71.2} \\
\bottomrule
\end{tabular}
\caption{
Accuracy (\%) on first 250 EgoLifeQA questions. 
Rows vary memory representation visual evidence; columns vary with unified multimodal evidence.
}
\label{tab:memory-unification}
\end{table}

\paragraph{Construction-time unification mitigates retrieve-then-align bottlenecks.} 
Table~\ref{tab:memory-unification} and Figure~\ref{fig:analysis_memory_alignment} also show that construction-time unification consistently improves over retrieval-time fusion. 
Retrieval-time fusion follows a retrieve-then-align pattern: each modality is retrieved independently, so evidence that is weakly relevant in isolation may be discarded before it can be combined with complementary cues from other modalities. 
In contrast, construction-time unification aligns captions, transcripts, and structured visual fields into shared event-level retrieval units before indexing. 
This makes retrieved evidence more semantically coherent, temporally grounded, and generation-ready. 
It also preserves evidence whose relevance emerges only after cross-modal binding, such as an object mentioned visually but disambiguated by nearby dialogue, or an action described in text but grounded by keyframes.
The gain is largest for structured evidence (+3.2\%), suggesting that explicit fields such as entities and actions provide useful anchors for cross-modal alignment. 
Together, these results support our central claim: effective multimodal memory should unify heterogeneous evidence during memory construction, producing event-level retrieval units that are directly usable for grounded and attributable long-video QA.

\paragraph{Keyframes serve as lightweight verification after memory retrieval.}
We further study how many keyframes should be included in the query-specific evidence view after relevant event cells have been selected. 
This analysis asks whether event-level textual and structured memory is sufficient, or whether answer generation still benefits from a small amount of visual evidence for verification. 
As shown in Figure~\ref{fig:analysis_keyframes}, overall accuracy improves from 63.2\% without keyframes to 66.0\% with three keyframes. 
The gains are larger for TaskMaster and RelationMap, where visual evidence helps verify task states, participant identities, and interaction context. 
By contrast, HabitInsight benefits less consistently because it depends more on repeated long-term behavioral patterns than on local visual verification. 
Thus, EM\textsuperscript{2}Mem uses compact event-centric multimodal memory as primary evidence and keyframes only for lightweight grounding.

\vspace{2mm}
\begin{table}[!htbp]
\centering
\small
\setlength{\tabcolsep}{4.5pt}
\renewcommand{\arraystretch}{1.15}
\begin{tabular*}{\columnwidth}{@{\extracolsep{\fill}} lccccc @{}}
\toprule
\multirow{2}{*}{\textbf{Type}}
& \multicolumn{2}{c}{\textbf{WorldMM}}
& \multicolumn{2}{c}{\textbf{EM\textsuperscript{2}Mem}}
& \multirow{2}{*}{\textbf{$\Delta$}} \\
\cmidrule(lr){2-3}
\cmidrule(lr){4-5}
& \textbf{1R} & \textbf{5R}
& \textbf{Top-1} & \textbf{Top-5}
& \\
\midrule
Overall & 15.6 & 23.8 & 23.0 & \textbf{30.8} & +7.0 \\
\midrule
Ent.  & 13.6 & 21.6 & 20.0 & \textbf{27.2} & +5.6 \\
EvR.  & 19.8 & \textbf{24.6} & 19.0 & 23.8 & -0.8 \\
Hab.  & 8.2  & 23.0 & 27.9 & \textbf{39.3} & +16.3 \\
Rel.  & 16.0 & 24.0 & 21.6 & \textbf{31.2} & +7.2 \\
Task. & 17.5 & 27.0 & 34.9 & \textbf{42.9} & +15.9 \\
\bottomrule
\end{tabular*}
\caption{
Strict 30-second event-level evidence recall (\%). 
$\Delta$ denotes the absolute gain of EM\textsuperscript{2}Mem Top-5 over WorldMM 5R.
}
\label{tab:compact_recall}
\end{table}

\paragraph{Event anchors enable single-pass evidence localization.}
We evaluate strict 30-second event-level recall, where a retrieval is correct only if the retrieved event anchor matches the annotated evidence segment. 
As shown in Table~\ref{tab:compact_recall} and Figure~\ref{fig:analysis_selector_budget}, EM\textsuperscript{2}Mem achieves 30.8\% Top-5 recall, outperforming WorldMM after five iterative retrieval rounds by 7.0 points. 
Its Top-1 recall also reaches 23.0\%, close to WorldMM 5R at 23.8\%, indicating that event-indexed memory can localize relevant evidence with a single-pass retrieval procedure. 
The gains are largest on HabitInsight and TaskMaster, where evidence often depends on long-term patterns or task states, while EventRecall remains slightly lower, suggesting that localized recall questions may still benefit from iterative caption-level retrieval.
We provide additional analysis of selector budgets and a qualitative case study in Appendix~\ref{app:selector-budget} and Appendix~\ref{app:case-study}.

\section{Related Work}

\paragraph{Long Video Understanding.}
Video-language research has progressed from video-text pretraining and temporal-aware representation learning to video LLMs that connect visual encoders with LLMs for instruction following and long-form understanding \cite{VideoBert, VideoMAE, InternVideo2, Video-LLaMA, Video-ChatGPT, Chat-UniVi, LLaVA-OneVision}. 
Recent methods extend video LLMs to longer contexts through visual token compression, sparse or hierarchical memory, long-context adaptation, and temporal reasoning mechanisms \cite{LongVA, LongVU, VideoChat-Flash, TimeChat, Time-r1, Video-RTS, Video-Thinker}. 
However, as videos scale from minutes to hours or days, relevant evidence becomes sparse, temporally distant, and distributed across modalities, motivating explicit mechanisms for organizing and retrieving video information.

\paragraph{Multimodal Memory for LLMs.} 
Memory- and retrieval-augmented systems address long-video QA by storing captions, transcripts, keyframes, OCR, clip embeddings, or graph-structured indices and retrieving question-relevant evidence before generation \cite{LightRAG, HippoRAG, Video-RAG__Visually-aligned, VideoRAG__Retrieval-Augmented, LangRepo}. 
Closely related egocentric and agentic systems construct hierarchical memories, episode summaries, or adaptive retrieval procedures for long-range reasoning \cite{EgoLife, HippoMM, VideoAgent, M3-Agent, Ego-R1, WorldMM, VideoARM}. 
We provide an extended related work in Appendix~\ref{app:extended_related_work}.

\section{Conclusion}

We presented EM\textsuperscript{2}Mem, an event-centric multimodal memory framework that organizes heterogeneous evidence under shared event anchors.
By retrieving aligned multimodal events rather than isolated fragments, EM\textsuperscript{2}Mem improves accuracy, evidence localization, and inference efficiency for scalable long-video question answering.



\section*{Limitations}
EM\textsuperscript{2}Mem has several limitations. 
Its structured memory design trades visual fidelity for searchability: converting visual evidence into textual fields makes it compatible with event-level retrieval, but may lose fine-grained pixel details such as small objects, colors, layouts, and subtle visual states. 
Future work may explore coarse-to-fine multimodal memory, where structured fields support retrieval while finer visual evidence is preserved for verification.
The align-then-retrieve paradigm is also not fully complete. 
Although multimodal evidence is aligned under event anchors before retrieval, the final answer stage still relies on selected keyframes for visually detailed reasoning. 
This leaves part of cross-modal alignment to inference time and may introduce modality bias.
EM\textsuperscript{2}Mem also depends on upstream MLLMs and visual tools, so errors in captioning, object extraction, or metadata normalization may propagate into event cells and graphs. 
In addition, the framework shifts computation from inference to memory construction, making it more suitable for videos processed once and queried repeatedly than for real-time scenarios.

\section*{Ethical Statement}
This work studies long-video question answering with multimodal memory, including evaluations on egocentric daily-life benchmarks. Such videos may contain privacy-sensitive information, including personal routines, indoor environments, object usage, social interactions, and potentially identifying visual or textual cues. 
We do not collect new human-subject data, and all experiments are conducted on publicly released benchmark datasets under their intended research settings. 
We do not attempt to identify individuals, infer protected attributes, or disclose raw video content or personally identifiable information; all results are reported in aggregate. 
Since EM\textsuperscript{2}Mem organizes events, entities, routines, and relations into structured memory, misuse on unconstrained personal videos could increase risks of privacy leakage, behavioral profiling, or surveillance-like applications. 
Therefore, practical deployment of such memory systems should require informed consent, access control, data minimization, memory deletion or expiration mechanisms, and safeguards against using inferred semantic facts for high-stakes decisions. 
We also note that egocentric video benchmarks may reflect demographic and environmental biases, so performance should not be assumed to generalize uniformly across populations or deployment contexts.

\section*{Acknowledgements}
We would like to express our sincere gratitude to the anonymous reviewers for their thoughtful and constructive feedback. This work was supported by the New Generation Artificial Intelligence-National Science and Technology Major Project (2025ZD0122802), National Natural Science Foundation of China (No.62676362, No.NSFCU23B2055, No.NSFCU19B2027), the Fundamental Research Funds for the Central Universities (226-2023-00138), the Yongjiang Talent Introduction Programme (2021A-156-G), and the Information Technology Center and State Key Lab of CAD\&CG, Zhejiang University. This work was sponsored by CCF-Lenovo Blue Ocean Research Fund.

\bibliography{custom}

\appendix

\clearpage
\section*{Appendix}
\section{Dataset Details}
\label{app:dataset}
This section provides additional details about the benchmark datasets used in the experiments.
As shown in Table~\ref{tab:dataset_summary}, we evaluate EM\textsuperscript{2}Mem on three long-video question answering benchmarks, covering both ultra-long egocentric daily-life videos and general long-form videos. 
EgoLifeQA and Ego-R1 Bench focus on first-person daily-life reasoning over hour-level video histories, while Video-MME (L) evaluates general long-video understanding across diverse visual domains.
\begin{table}[H]
\centering
\small
\setlength{\tabcolsep}{3.5pt}
\renewcommand{\arraystretch}{0.95}
\begin{tabular}{@{}lccc@{}}
\toprule
\textbf{Dataset} & \textbf{\#Queries} & \textbf{Domain} & \textbf{Avg. Video Length} \\
\midrule
EgoLifeQA      & 500 & Egocentric & 44.3h \\
Ego-R1 Bench   & 300 & Egocentric & 44.3h \\
Video-MME (L)  & 900 & General & 0.69h \\
\bottomrule
\end{tabular}
\caption{Summary of benchmark datasets used in our experiments.}
\label{tab:dataset_summary}
\end{table}

\subsection{EgoLifeQA}
EgoLifeQA~\cite{EgoLife} is a long-context egocentric video question answering benchmark built on week-long first-person daily-life recordings. 
It is designed to evaluate whether a model can serve as a personalized life assistant by retrieving and reasoning over sparse evidence distributed across long video histories. 
Following prior work, we evaluate on the A1\_JAKE subject stream, which contains 44.3 hours of video and 500 multiple-choice questions.

The benchmark covers five types of memory-intensive queries. 
\textit{EntityLog} questions require tracking objects, entities, and their states; 
\textit{EventRecall} focuses on recalling specific past events and their temporal context; 
\textit{HabitInsight} requires identifying recurring behaviors or long-term preferences; 
\textit{RelationMap} evaluates understanding of social interactions and relationships; 
and \textit{TaskMaster} asks about ongoing plans or pending tasks. 
This dataset is therefore well aligned with our goal of evaluating whether structured event memory can support object-centric, event-centric, and long-term semantic reasoning in ultra-long egocentric videos.

\subsection{Ego-R1 Bench}
Ego-R1 Bench~\cite{Ego-R1} evaluates ultra-long egocentric video reasoning with 300 multiple-choice questions. 
It shares the daily-life video setting with EgoLifeQA, but emphasizes multi-step evidence gathering and tool-augmented reasoning over long video histories. 
To enable consistent analysis across egocentric benchmarks, we map its original query types to the five EgoLifeQA categories, as shown in Table~\ref{tab:egor1_mapping}.
\begin{table}[H]
\centering
\small
\setlength{\tabcolsep}{3pt}
\renewcommand{\arraystretch}{0.95}
\begin{tabularx}{\columnwidth}{@{}lX@{}}
\toprule
\textbf{Category} & \textbf{Ego-R1 Query Types} \\
\midrule
EntityLog & EntityLog, FoodLog, HealthLog, TechLog \\
EventRecall & EventRecall, Event Recollection, Event Memory \\
HabitInsight & HabitInsight, Behavior Habit(s) \\
RelationMap & RelationMap, Interpersonal Relationships \\
TaskMaster & TaskMaster, Future Plan(s) \\
\bottomrule
\end{tabularx}
\caption{Mapping Ego-R1 Bench query types to the EgoLifeQA category taxonomy.}
\label{tab:egor1_mapping}
\end{table}

\subsection{Video-MME}
Video-MME~\cite{Video-MME} is a comprehensive benchmark for evaluating MLLMs on general video understanding, covering diverse domains, temporal durations, and multimodal inputs. 
Different from EgoLifeQA and Ego-R1 Bench, which focus on egocentric daily-life memory, Video-MME provides an open-domain evaluation setting for long-form video comprehension. 
In our experiments, we use only the long-video subset, denoted as Video-MME (L), which contains videos longer than 30 minutes and 900 multiple-choice questions. 
We adopt the benchmark's original task types.

\subsection{License}
We use only publicly available benchmarks released by their original authors. 
All datasets are used for academic evaluation under their corresponding licenses and terms of use. 
We do not redistribute original videos, annotations, or benchmark files, and users should obtain the datasets from the official sources and comply with the original licenses. 
For benchmarks with non-commercial or academic-use restrictions, such as Video-MME and EgoLife, our use is limited to research purposes; for permissively licensed datasets such as Ego-R1-Data, we follow their license terms.

\section{Implementation Details}
\label{app:implementation_details}
This section provides additional details regarding the baseline implementations, the configuration of EM\textsuperscript{2}Mem, and the prompts employed in our experiments.

\subsection{Baseline Setup}
\label{app:baseline-setup}

We compare EM\textsuperscript{2}Mem with a broad set of baselines covering four categories.

\paragraph{Base MLLMs.}
We include GPT-5 \cite{GPT-5}, Gemini 2.5 Pro \cite{Gemini-2.5-pro}, and Qwen3-VL-8B-Instruct \cite{Qwen3-VL}. These models are evaluated as general-purpose multimodal LLMs without an external long-video memory module.

\paragraph{Long-video MLLMs.}
We compare with VideoChat-Flash \cite{VideoChat-Flash}, Time-R1 \cite{Time-r1}, and Video-RTS \cite{Video-RTS}, which process sampled visual inputs within their respective context or frame-budget constraints. 
These baselines test whether stronger long-video input processing alone is sufficient for long-horizon QA.

\paragraph{RAG-based methods.}
We include LightRAG \cite{LightRAG}, HippoRAG \cite{HippoRAG}, and Video-RAG \cite{Video-RAG__Visually-aligned}. 
LightRAG and HippoRAG retrieve from text-form video evidence such as captions or summaries, while Video-RAG retrieves relevant video clips before answer generation. 
These methods represent retrieve-then-generate pipelines without event-centric memory construction.

\paragraph{Memory-based long-video reasoning.}
We compare against EgoRAG \cite{EgoLife}, Ego-R1 \cite{Ego-R1}, HippoMM \cite{HippoMM}, M3-Agent \cite{M3-Agent}, and WorldMM \cite{WorldMM}. 
These methods maintain external memories or agentic retrieval procedures for long-video reasoning, making them the closest family of baselines to EM\textsuperscript{2}Mem.

\paragraph{Reported and reproduced results.}
For baselines other than WorldMM, we use the results reported in WorldMM \cite{WorldMM} on the overlapping benchmarks, including EgoLifeQA, Ego-R1 Bench, and Video-MME (L), to keep dataset splits, metrics, and evaluation protocols consistent. 
Since WorldMM is the strongest and most related memory-based baseline, we additionally reproduce it under the same evaluation setting as EM\textsuperscript{2}Mem and denote the reproduced version as WorldMM$^\dagger$.

\paragraph{WorldMM$^\dagger$ reproduction.}
We follow the original WorldMM configuration wherever possible. To ensure a controlled comparison with EM\textsuperscript{2}Mem, WorldMM$^\dagger$ uses the same backbone as our method when applicable: GPT-5-mini-2025-08-07 \cite{GPT-5} is used during memory construction, including episodic and semantic memory construction and LLM-based triplet consolidation, while GPT-5-2025-08-07 is used as the retrieval agent and final answer generator. 
This corresponds to the WorldMM-GPT setting in the original paper. For episodic memory, we use temporal scales of 30 seconds, 3 minutes, 10 minutes, and 1 hour for EgoLifeQA and Ego-R1 Bench, and 10 seconds, 30 seconds, 3 minutes, and 10 minutes for Video-MME (L). 
For semantic memory, triplets with a similarity score greater than 0.6 are consolidated using an LLM, and the top-10 most relevant triplets are retrieved during inference.
We keep the original visual memory design with both feature-based retrieval and timestamp-based frame access. 
The retrieval agent is allowed up to five iterations, and the remaining memory construction, retrieval, and response generation pipeline is kept unchanged.

\subsection{EM\texorpdfstring{\textsuperscript{2}}{²}Mem}

\paragraph{Memory Construction Details.} 
EM\textsuperscript{2}Mem is implemented as a training-free framework without updating model parameters. 
During memory construction, we use GPT-5-mini-2025-08-07 to convert event anchors into multimodal event records containing captions, transcripts, structured metadata, visual fields, timestamps, and provenance. 
We build temporal context views at 30 seconds, 3 minutes, 10 minutes, and 1 hour for EgoLifeQA and Ego-R1 Bench, and at 10 seconds, 30 seconds, 3 minutes, and 10 minutes for Video-MME (L). 
Episodic triplets are extracted from both local event records and temporal context views to construct the episodic graph. 
For semantic memory, we follow WorldMM~\cite{WorldMM}: triplets with similarity above 0.6 are consolidated with an LLM.

\paragraph{Inference Details.} 
During inference, EM\textsuperscript{2}Mem retrieves evidence through event anchors with \(K_E=5\) episodic candidates and \(K_S=8\) semantic facts. 
Each candidate event \(e_i\) is scored using the local multimodal event record \(R_i\), temporal context views \(C_i\), episodic graph evidence \(G_i\), and semantic facts \(S_i\). 
We assign weight \(1.00\) to the local event record, use scale weights \(0.65/0.45/0.30\) for 3-minute/10-minute/1-hour context views, and apply a one-hop graph expansion decay of \(0.60\). 
The LLM selector, implemented with GPT-5-2025-08-07, selects \(K_{\text{sel}}=5\) event anchors; for each selected anchor, we attach up to \(K_V=3\) keyframes for final answer generation with GPT-5. 
Experiments are run on two NVIDIA A100 40GB GPUs for embedding-index storage and retrieval, with 8 parallel workers during inference evaluation.

For reproducibility, we provide a repository containing our implementation,
prompts, configuration files, and evaluation scripts at
\url{https://github.com/zjunlp/LightMem}.

\subsection{Prompts}
\begin{figure*}[p]
    \centering
    \includegraphics[width=1\textwidth]{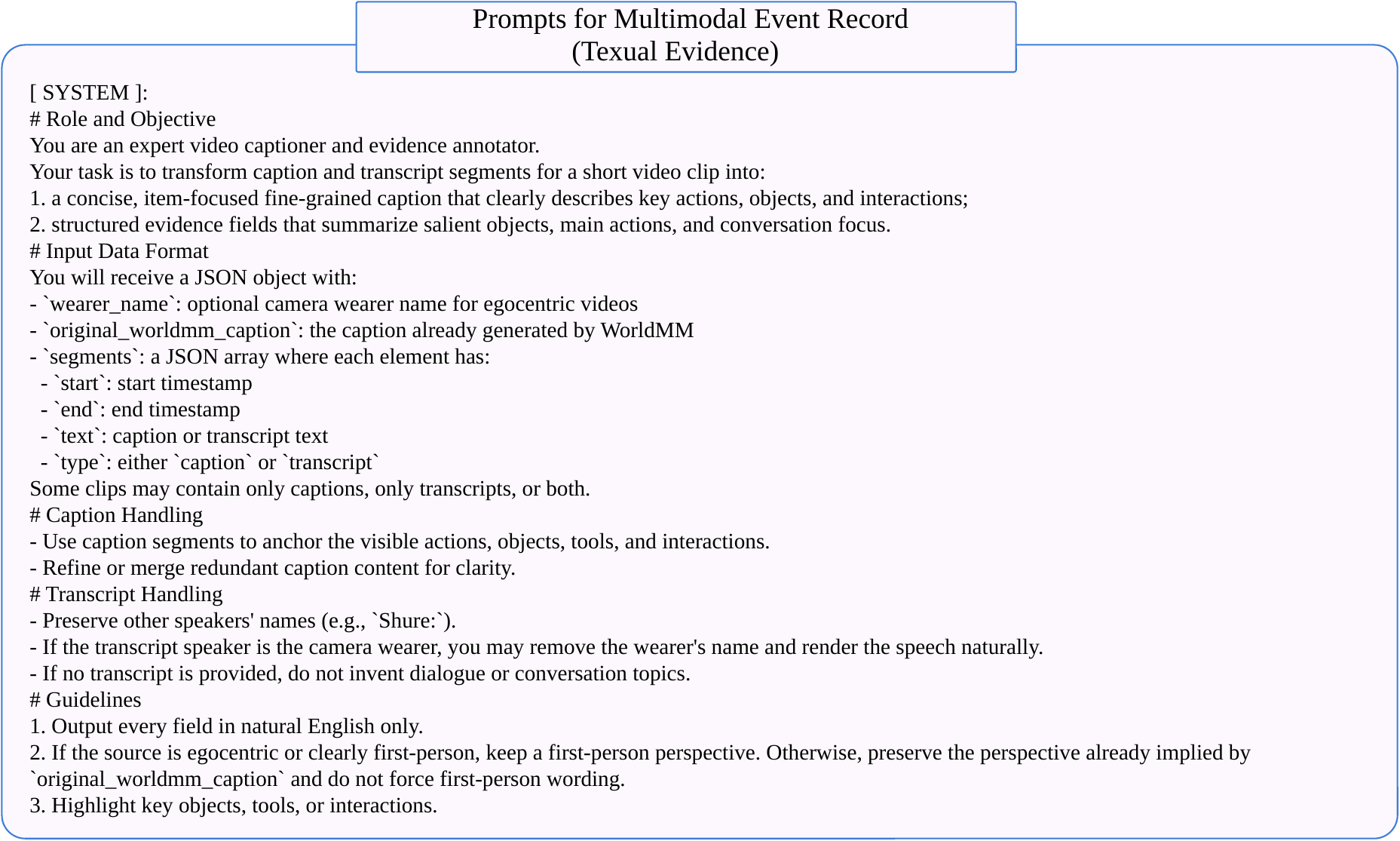}
    \caption{Prompt for constructing text-derived fields in Multimodal Event Records, part 1.}
    \label{fig:prompt_event_text_1}
\end{figure*}

\begin{figure*}[p]
    \centering
    \includegraphics[width=1\textwidth]{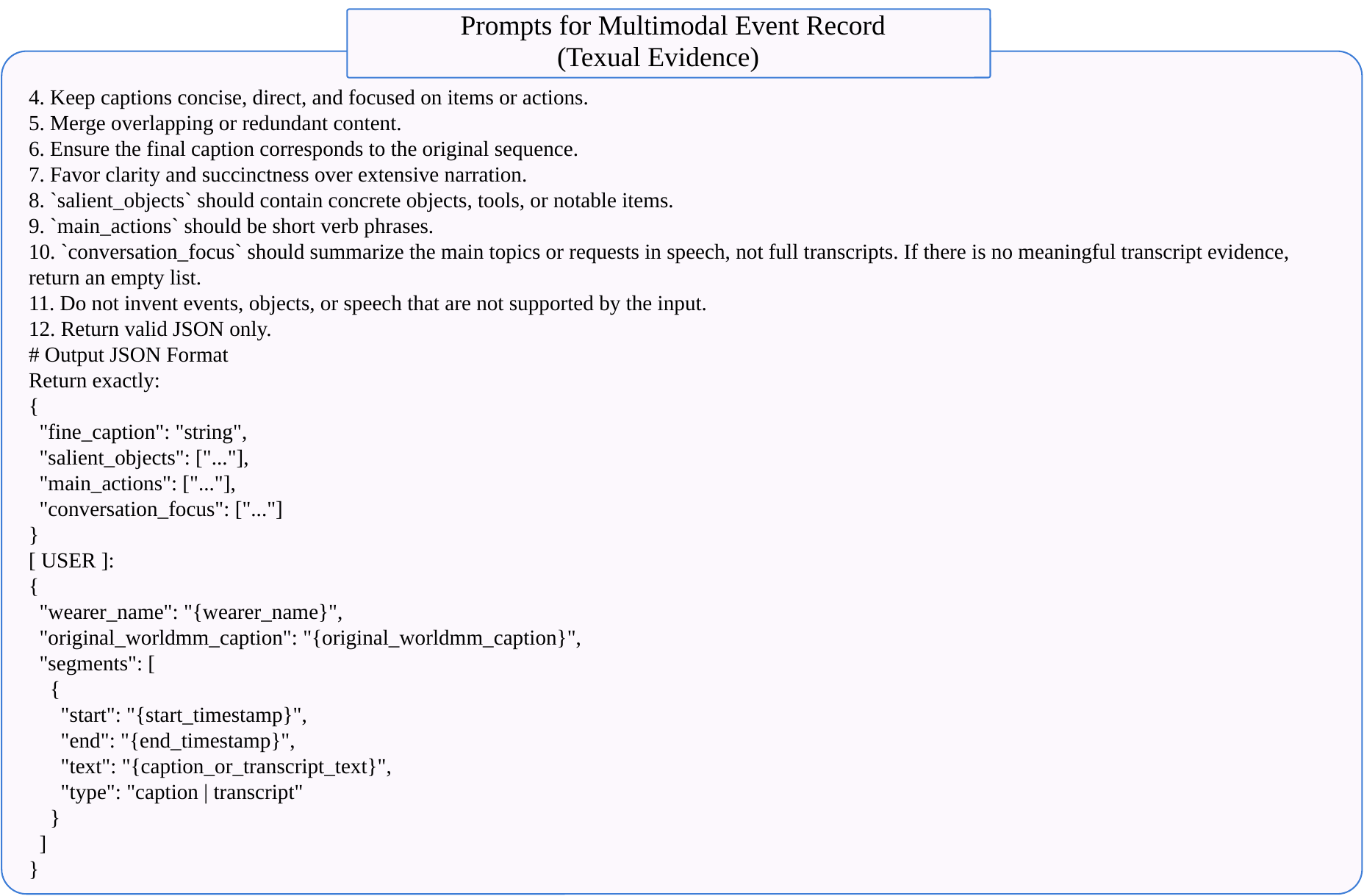}
    \caption{Prompt for constructing text-derived fields in Multimodal Event Records, part 2.}
    \label{fig:prompt_event_text_2}
\end{figure*}

\begin{figure*}[p]
    \centering
    \includegraphics[width=1\textwidth]{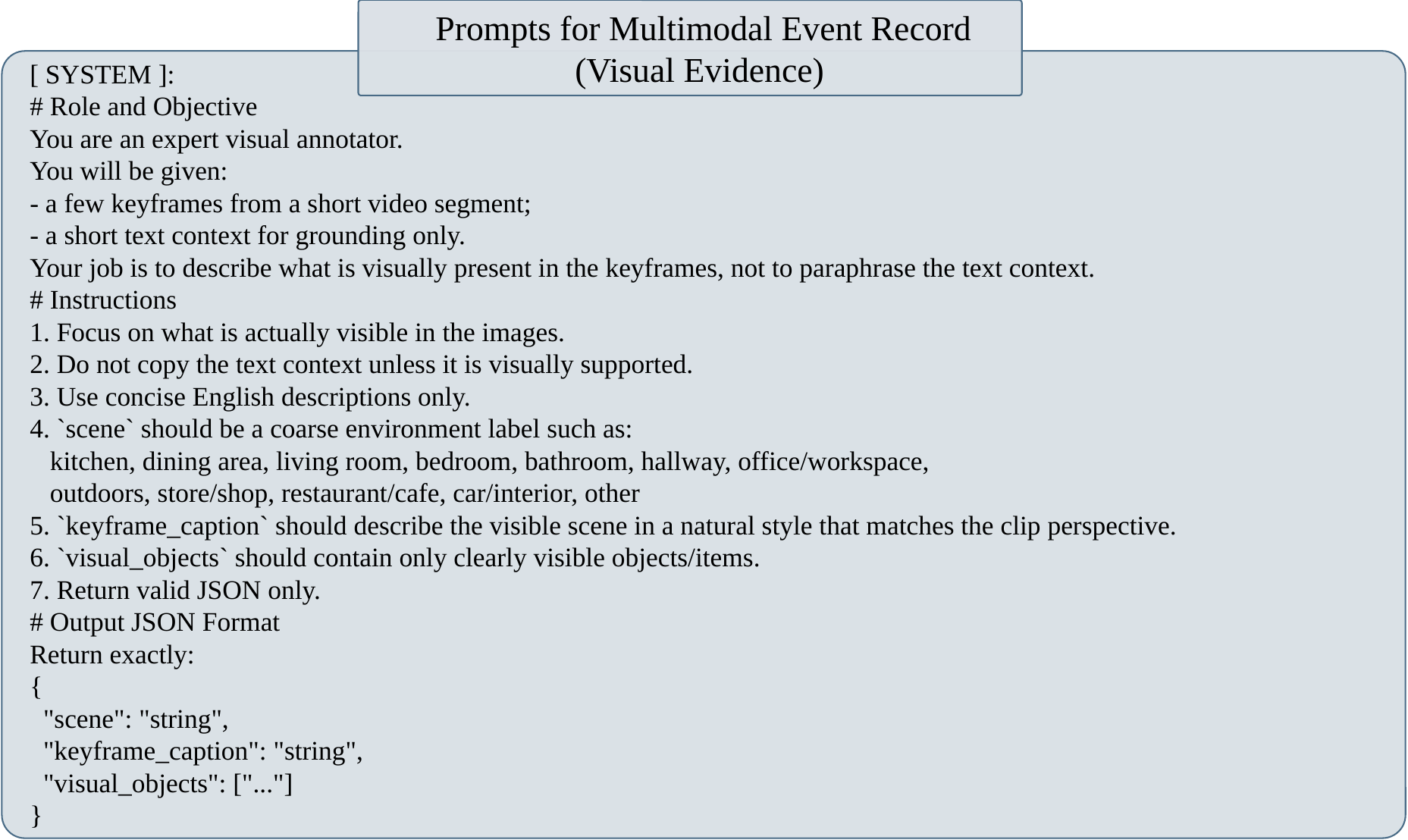}
    \caption{Prompt for constructing visual fields in Multimodal Event Records, part 1.}
    \label{fig:prompt_event_visual_1}
\end{figure*}

\begin{figure*}[p]
    \centering
    \includegraphics[width=1\textwidth]{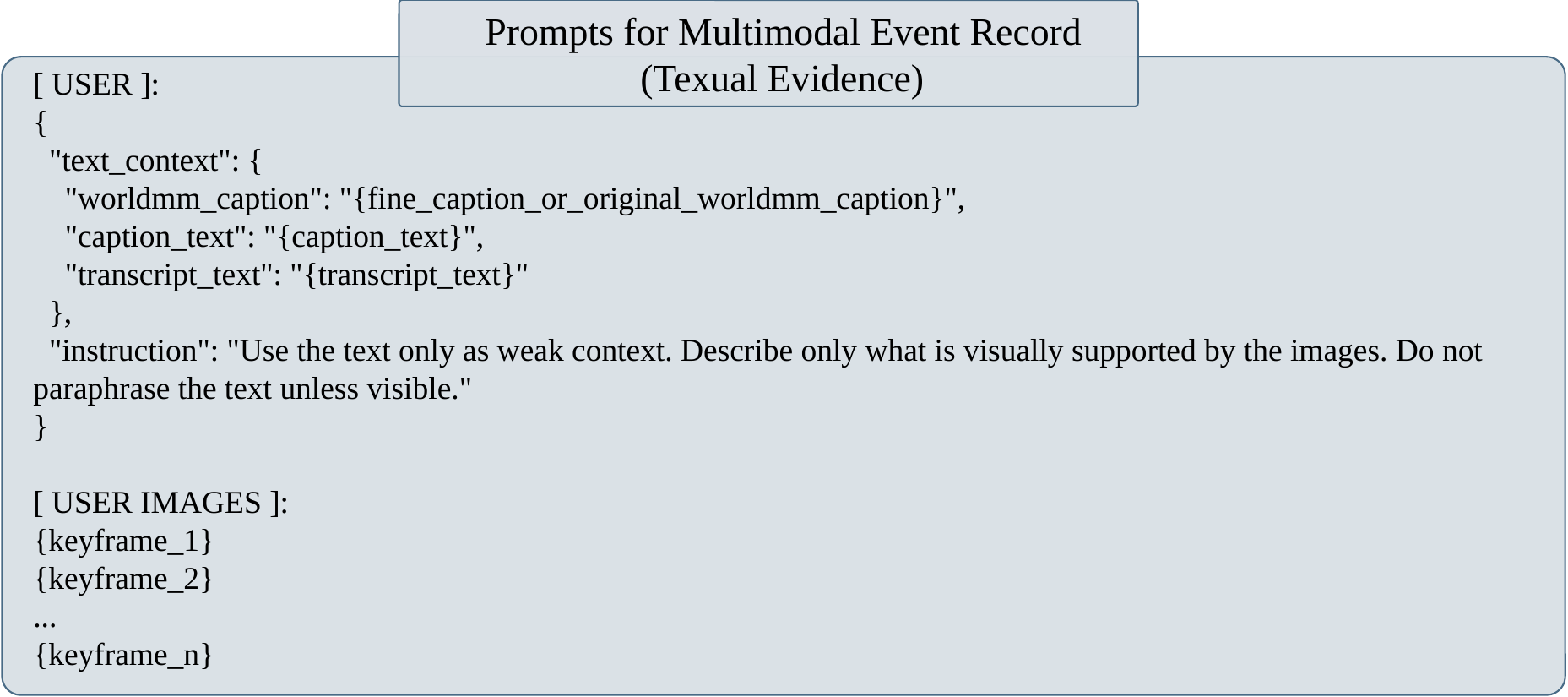}
    \caption{Prompt for constructing visual fields in Multimodal Event Records, part 2.}
    \label{fig:prompt_event_visual_2}
\end{figure*}

We report the prompt templates used in EM\textsuperscript{2}Mem for memory construction and inference. 
For memory construction, we design prompts for constructing Multimodal Event Records and Temporal Context Views. 
The text-side Multimodal Event Record prompt (Figures~\ref{fig:prompt_event_text_1} and~\ref{fig:prompt_event_text_2}) guides the model to refine short event captions and extract structured fields such as salient objects, main actions, and conversation focus from captions and transcripts. 
The visual-side Multimodal Event Record prompt (Figures~\ref{fig:prompt_event_visual_1} and~\ref{fig:prompt_event_visual_2}) instructs the model to annotate keyframes with visually grounded fields, including scenes, keyframe captions, and visual objects.

\begin{figure*}[p]
    \centering
    \includegraphics[width=1\textwidth]{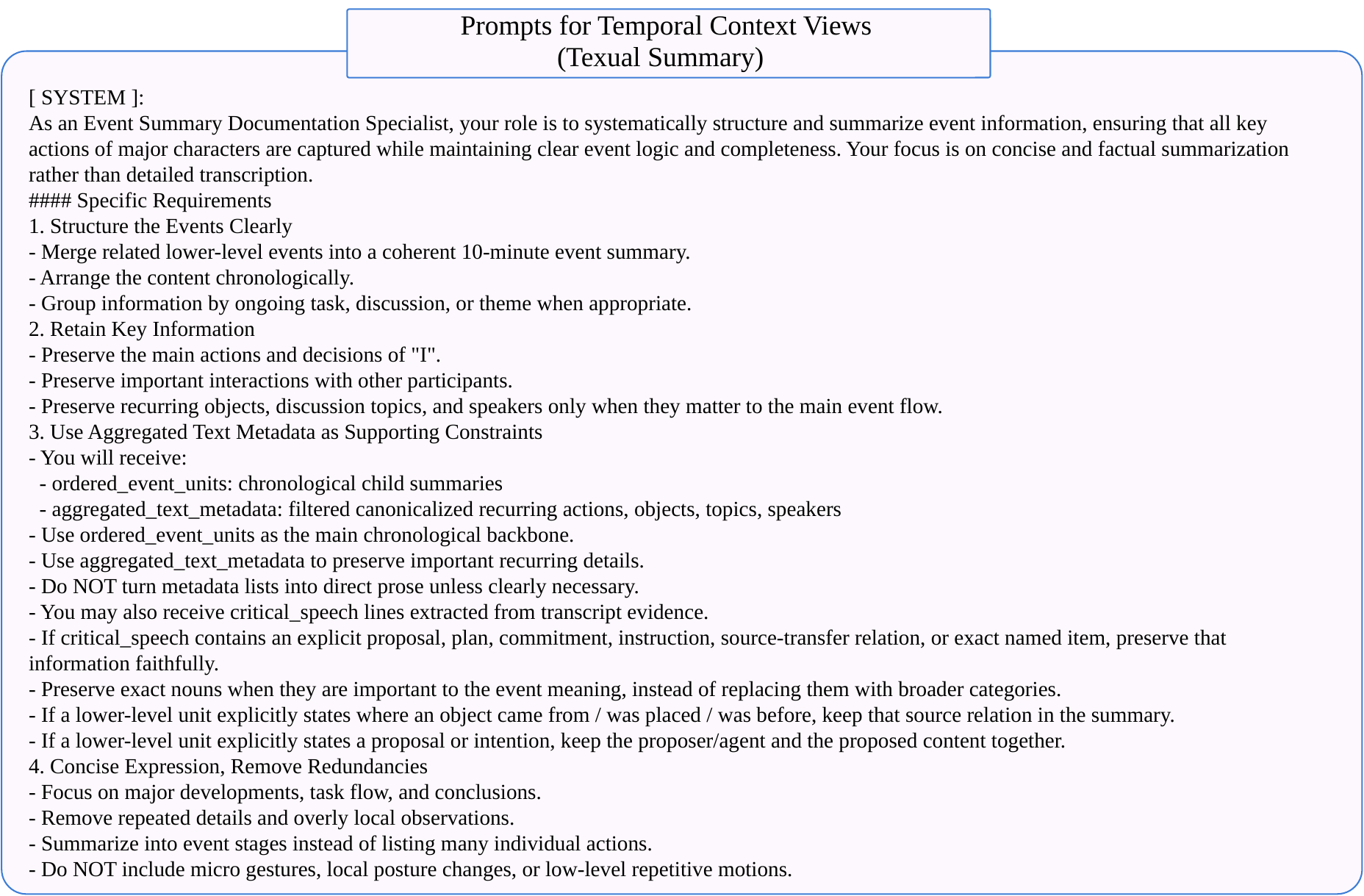}
    \caption{Prompt for constructing text summaries in Temporal Context Views, part 1.}
    \label{fig:prompt_multiscale_text_1}
\end{figure*}

\begin{figure*}[p]
    \centering
    \includegraphics[width=1\textwidth]{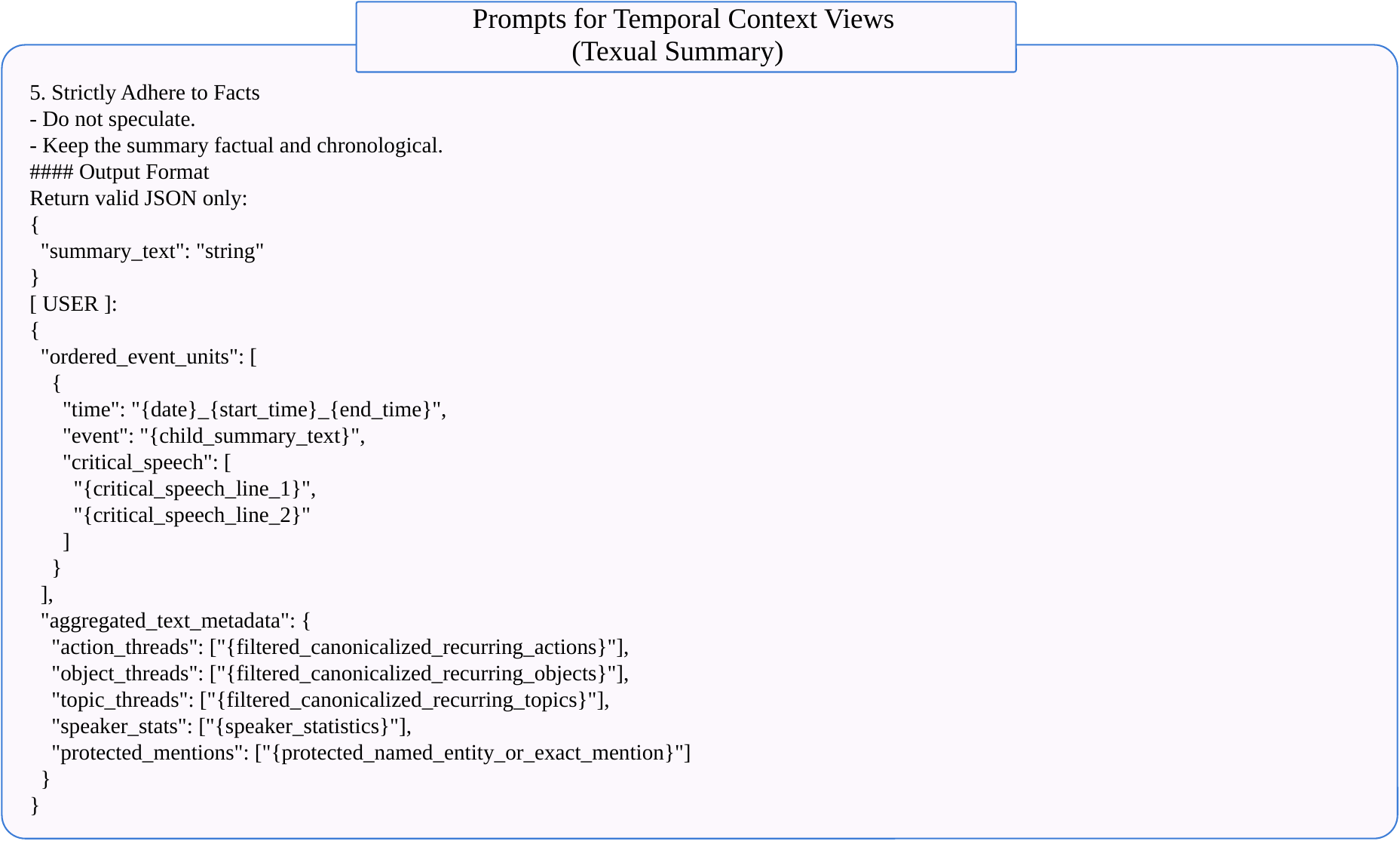}
    \caption{Prompt for constructing text summaries in Temporal Context Views, part 2.}
    \label{fig:prompt_multiscale_text_2}
\end{figure*}

\begin{figure*}[p]
    \centering
    \includegraphics[width=1\textwidth]{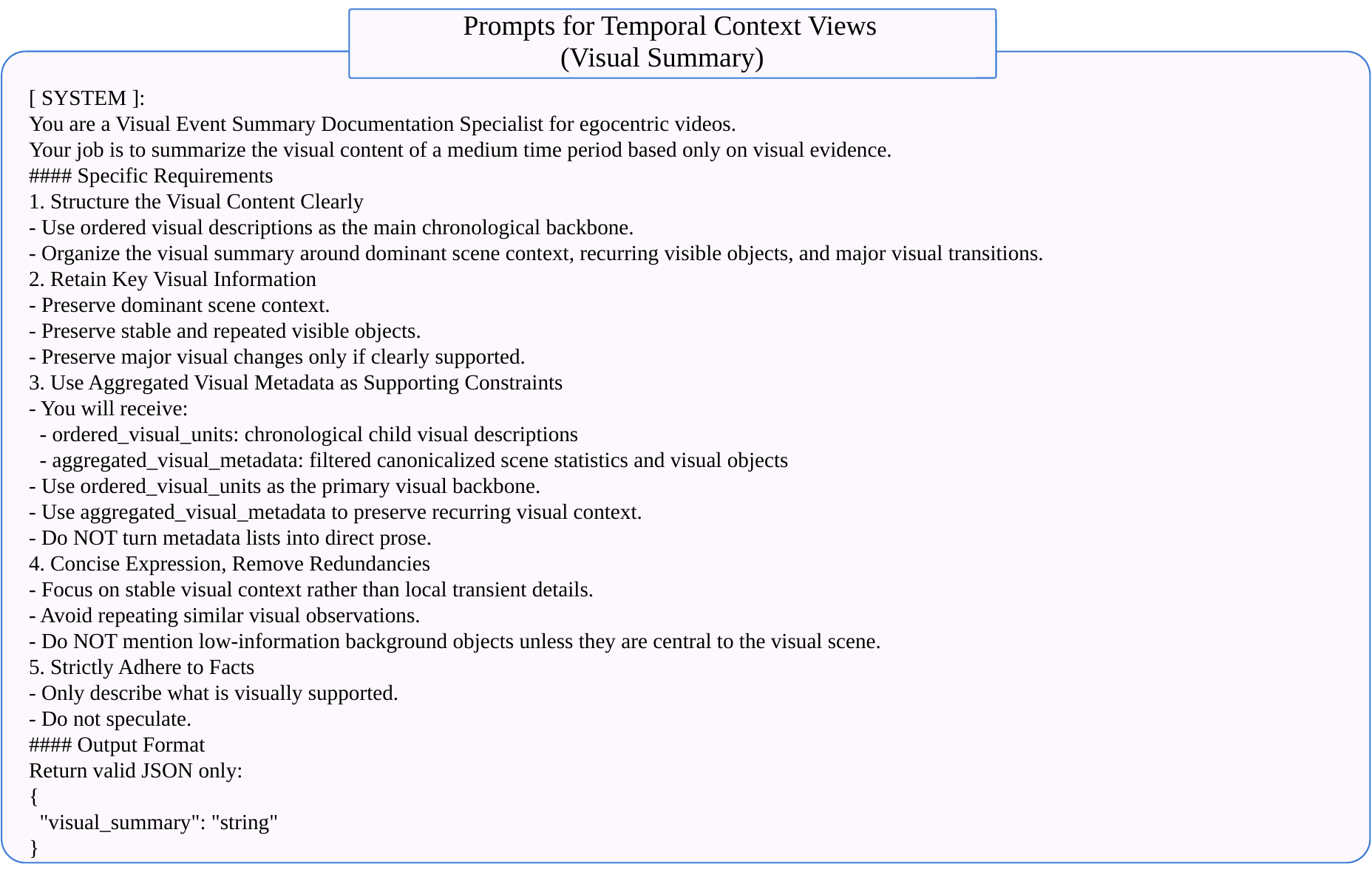}
    \caption{Prompt for constructing visual summaries in Temporal Context Views, part 1.}
    \label{fig:prompt_multiscale_visual_1}
\end{figure*}

\begin{figure*}[p]
    \centering
    \includegraphics[width=1\textwidth]{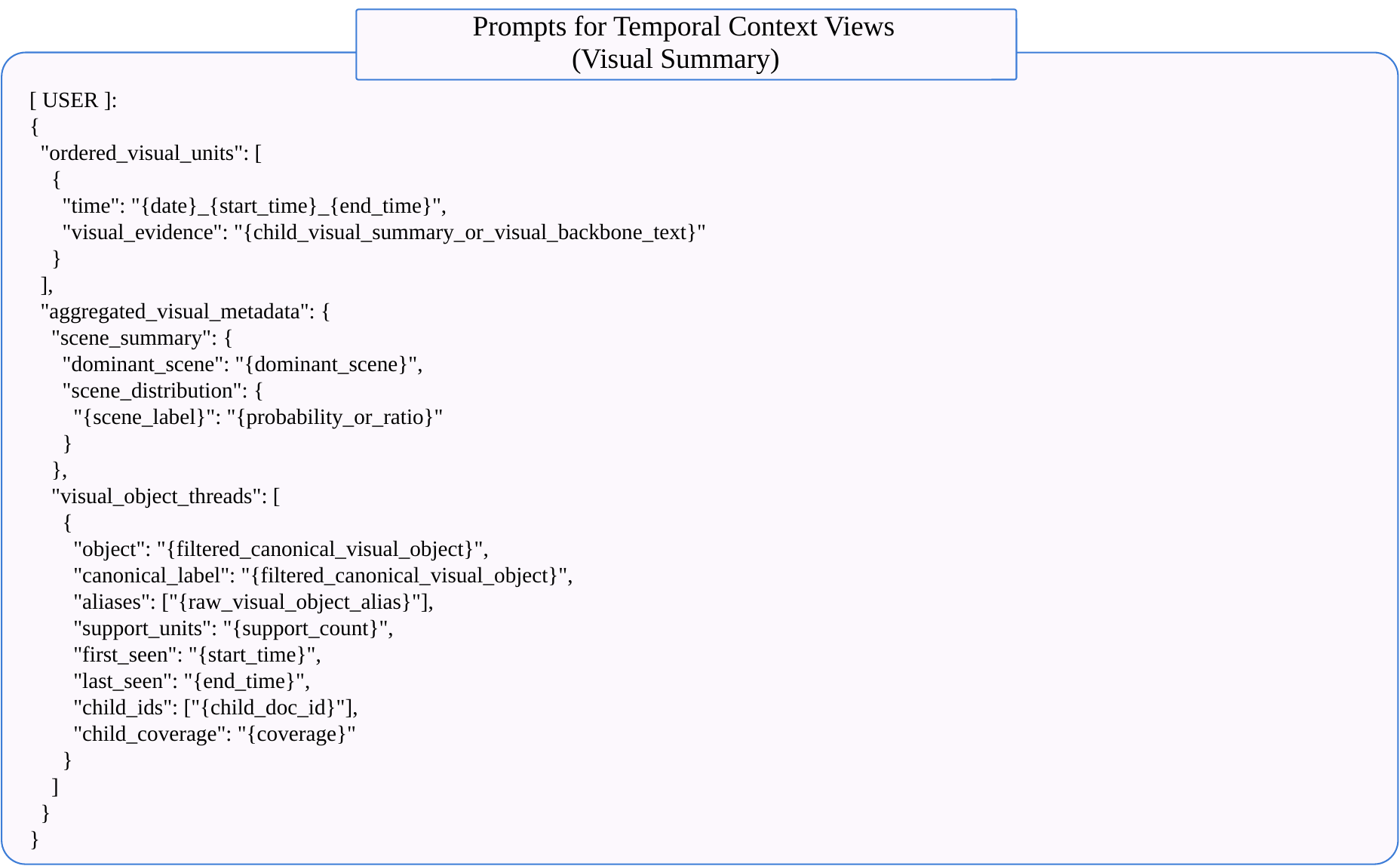}
    \caption{Prompt for constructing visual summaries in Temporal Context Views, part 2.}
    \label{fig:prompt_multiscale_visual_2}
\end{figure*}

For Temporal Context Views, the text summary prompt (Figures~\ref{fig:prompt_multiscale_text_1} and~\ref{fig:prompt_multiscale_text_2}) aggregates neighboring event records into coherent multi-scale textual summaries, while the visual summary prompt (Figures~\ref{fig:prompt_multiscale_visual_1} and~\ref{fig:prompt_multiscale_visual_2}) summarizes visual fields across temporal windows. 
Together, these prompts enable event-centered memories at multiple granularities.

\begin{figure*}[p]
    \centering
    \includegraphics[width=1\textwidth]{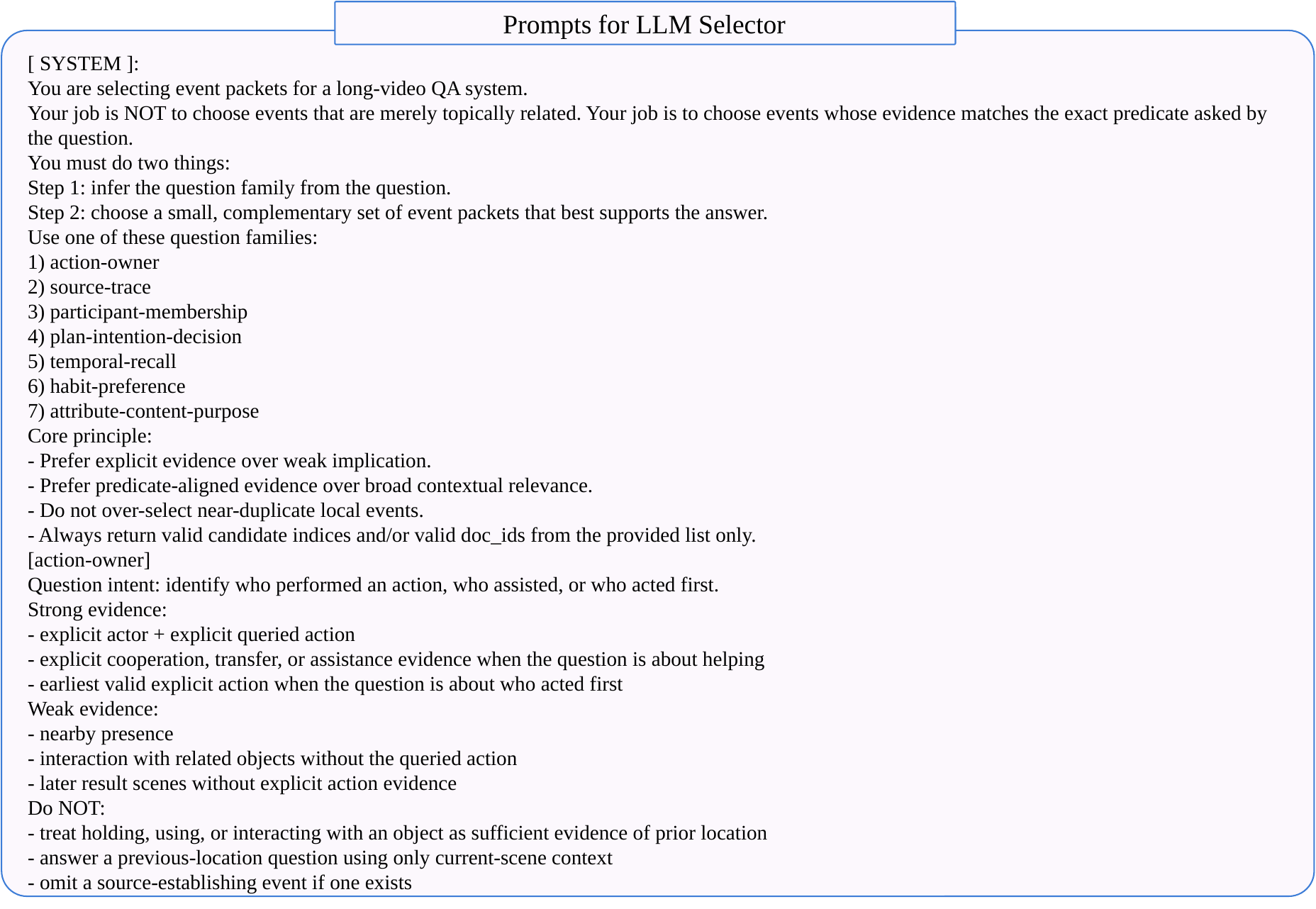}
    \caption{Prompt for the LLM Selector, part 1.}
    \label{fig:prompt_selector_1}
\end{figure*}

\begin{figure*}[p]
    \centering
    \includegraphics[width=1\textwidth]{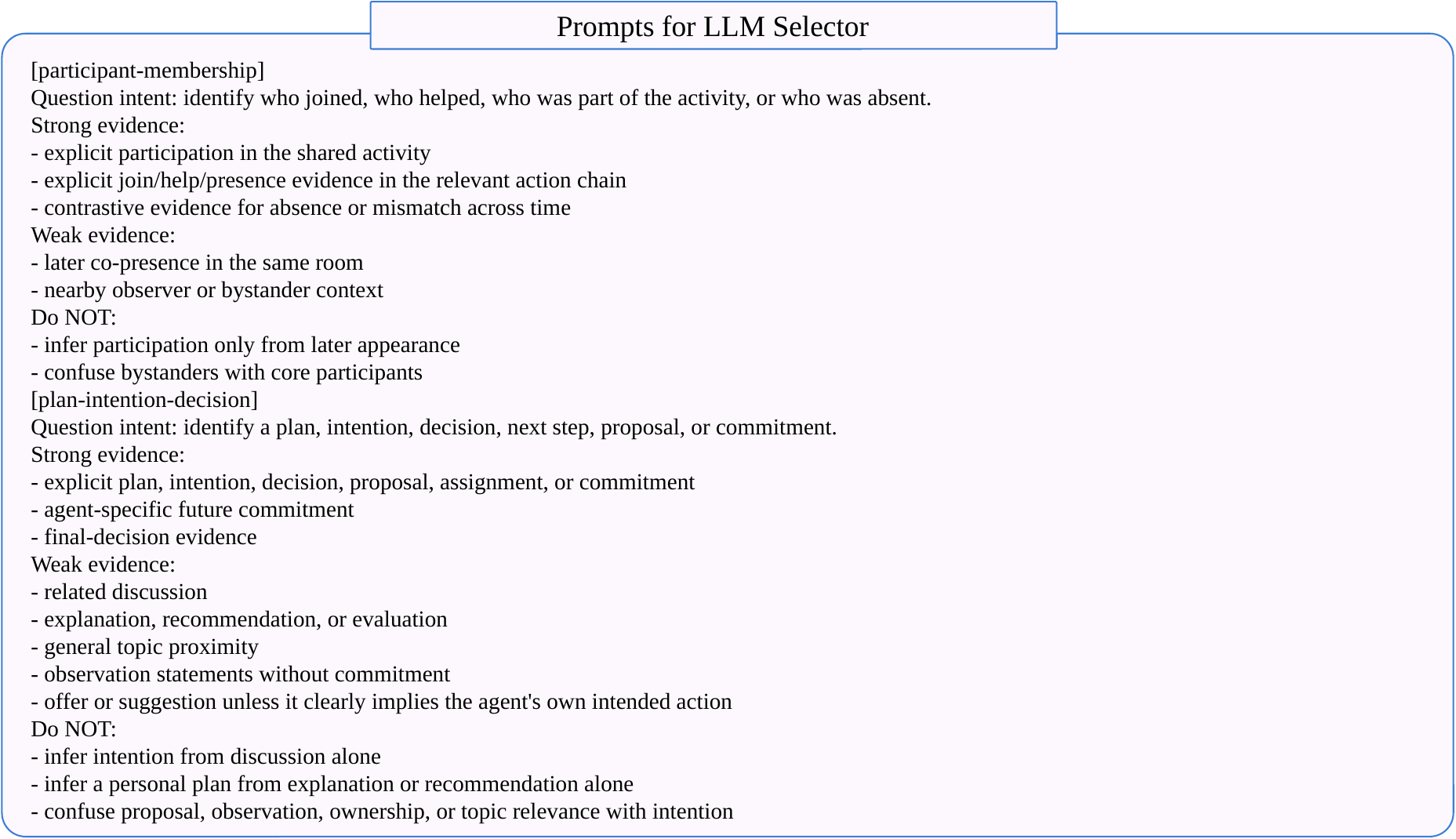}
    \caption{Prompt for the LLM Selector, part 2.}
    \label{fig:prompt_selector_2}
\end{figure*}

\begin{figure*}[p]
    \centering
    \includegraphics[width=1\textwidth]{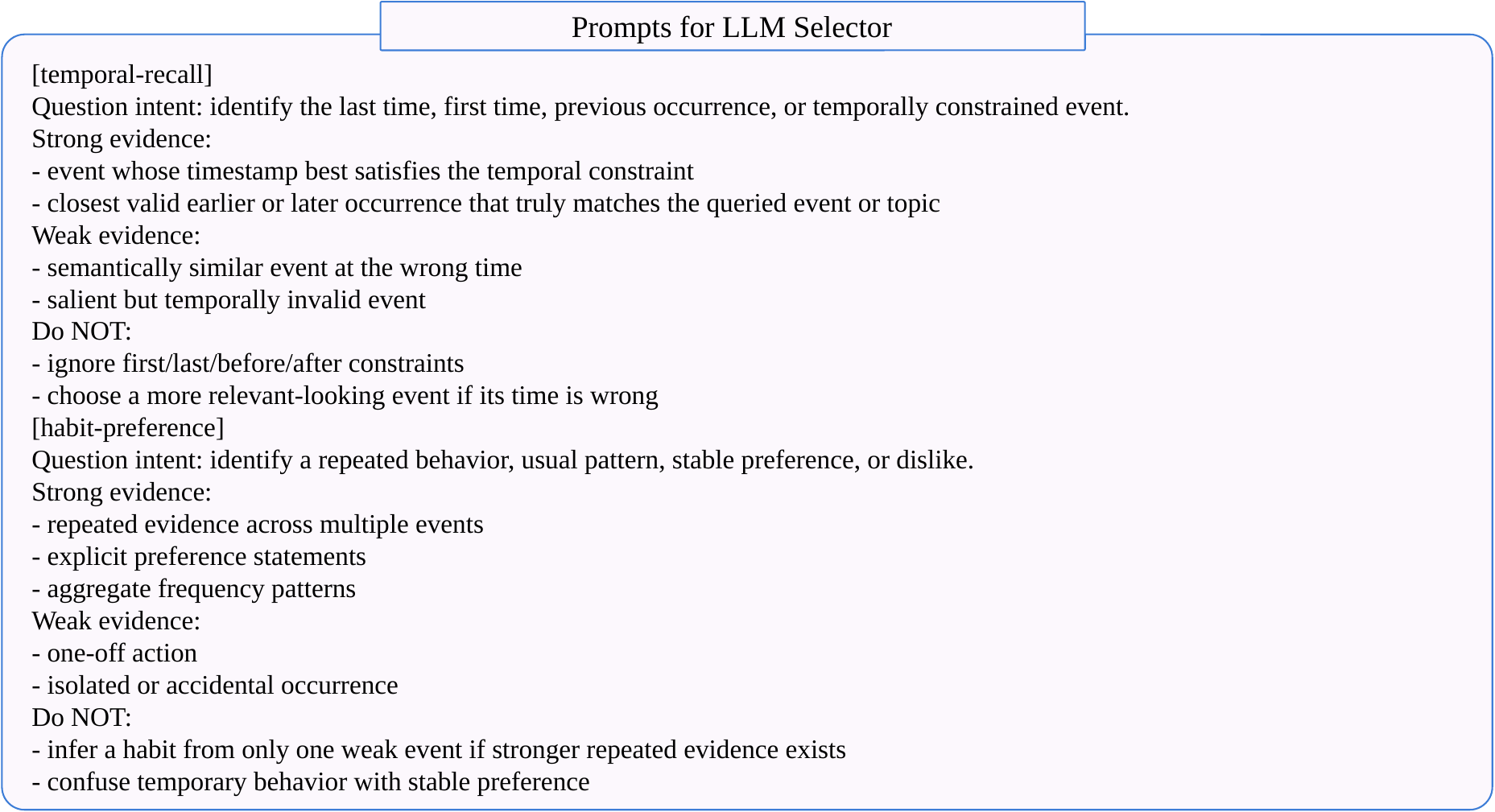}
    \caption{Prompt for the LLM Selector, part 3.}
    \label{fig:prompt_selector_3}
\end{figure*}

\begin{figure*}[p]
    \centering
    \includegraphics[width=1\textwidth]{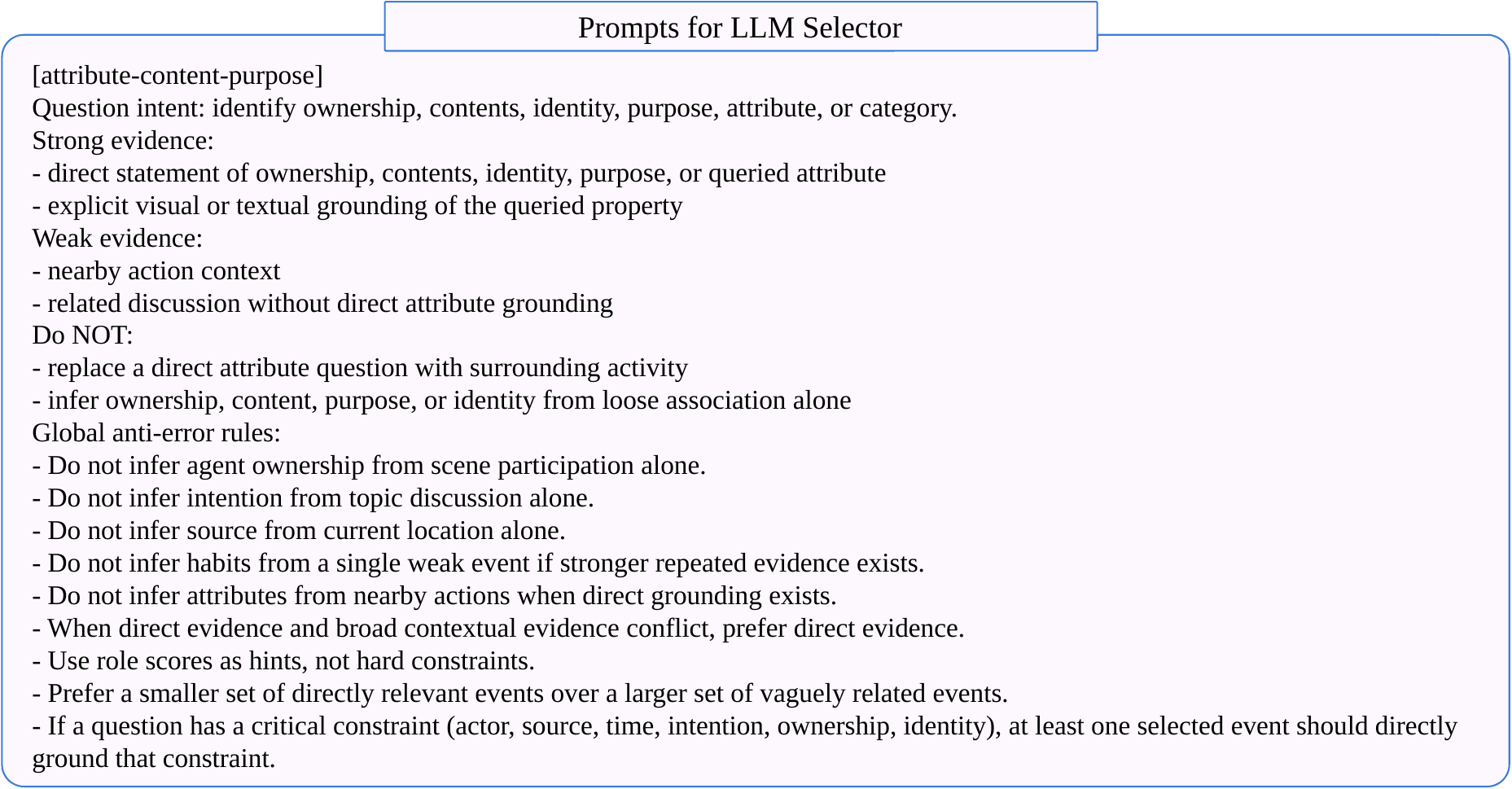}
    \caption{Prompt for the LLM Selector, part 4.}
    \label{fig:prompt_selector_4}
\end{figure*}

\begin{figure*}[p]
    \centering
    \includegraphics[width=1\textwidth]{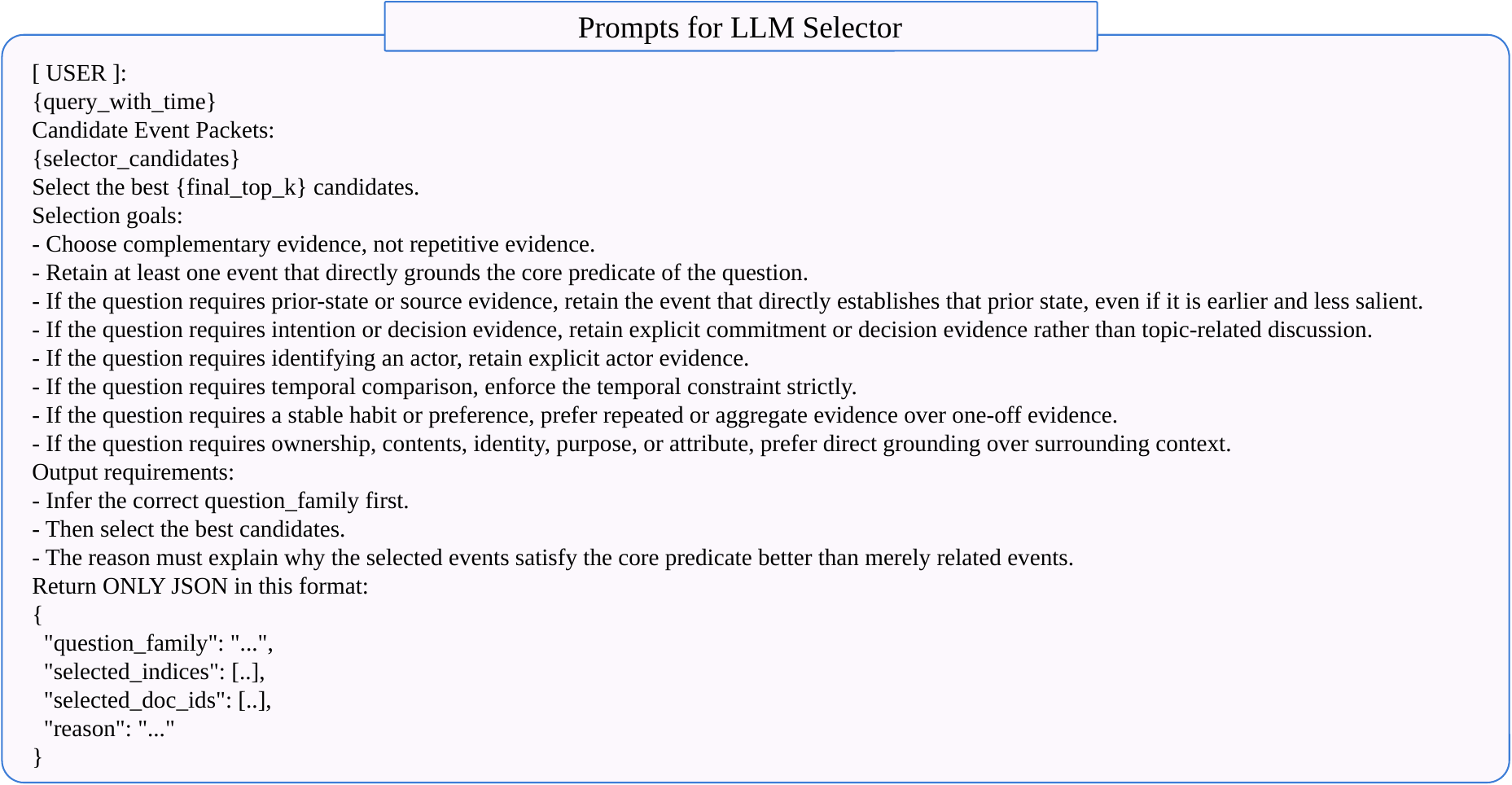}
    \caption{Prompt for the LLM Selector, part 5.}
    \label{fig:prompt_selector_5}
\end{figure*}

\begin{figure*}[p]
    \centering
    \includegraphics[width=1\textwidth]{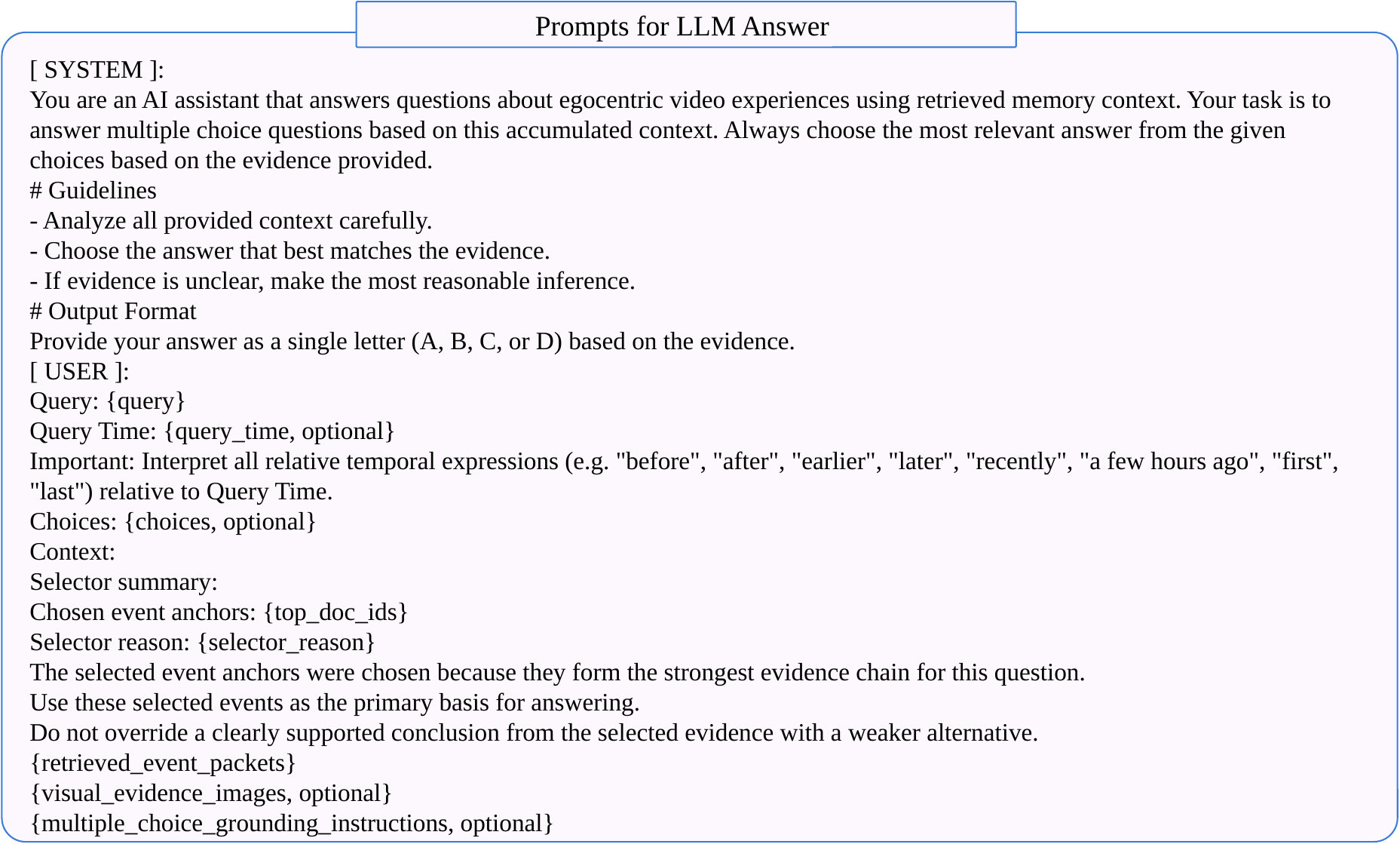}
    \caption{Prompt for the final Answer Model.}
    \label{fig:prompt_answer}
\end{figure*}

For inference, the LLM Selector prompt (Figures~\ref{fig:prompt_selector_1}--\ref{fig:prompt_selector_5}) selects the most relevant event anchors from retrieved event-centered candidates. 
The Answer Model prompt (Figure~\ref{fig:prompt_answer}) generates the final answer based on the selected structured evidence and attached visual keyframes.

\section{Additional Description on Experiments}

\subsection{Offline Memory Construction Cost}
\label{app:offline_cost}
Table~\ref{tab:memory-construction-cost} reports the token cost of offline memory construction. 
EM\textsuperscript{2}Mem follows an align-then-retrieve design that moves multimodal alignment and evidence organization from inference time to memory construction time. 
This introduces additional offline construction tokens, but organizes long-video evidence into compact event-centered memory units, enabling the inference stage to retrieve and reason over selected structured evidence instead of repeatedly aligning scattered modality-specific fragments. 
Since this cost is incurred once for each long-video memory, it can be amortized across repeated queries; as shown in Table~\ref{tab:efficiency}, the resulting event-indexed memory leads to a more lightweight inference process with lower per-query cost and higher efficiency.
\begin{table}[H]
\centering
\footnotesize
\setlength{\tabcolsep}{3.8pt}
\renewcommand{\arraystretch}{0.95}
\begin{tabular*}{\columnwidth}{@{\extracolsep{\fill}}lrrr@{}}
\toprule
\textbf{Method / Stage} 
& \textbf{Prompt} 
& \textbf{Completion} 
& \textbf{Total} \\
\midrule
WorldMM
& 23.26M 
& 37.29M 
& 60.55M \\
\midrule
\multicolumn{4}{@{}l}{\textit{EM\textsuperscript{2}Mem}} \\
Event Records
& 20.66M 
& 12.30M 
& 32.96M \\
Context Views
& 7.32M 
& 2.58M 
& 9.90M \\
Episodic Graph
& 9.07M 
& 16.85M 
& 25.92M \\
Semantic Memory
& 10.38M 
& 10.08M 
& 20.45M \\
EM\textsuperscript{2}Mem Total
& 47.43M 
& 41.81M 
& 89.23M \\
\bottomrule
\end{tabular*}
\caption{
Offline token cost of memory construction.
Prompt and completion denote model input and output tokens.
The construction cost is incurred once and amortized across repeated queries.
}
\label{tab:memory-construction-cost}
\end{table}

\subsection{End-to-End Cost and Amortization Analysis}
\label{app:e2e_cost}
\paragraph{End-to-end cost and amortization.}
EM\textsuperscript{2}Mem intentionally shifts part of the computation from query-time reasoning to offline memory construction.
This design is motivated by the observation that long-video memories are typically constructed once but queried multiple times.
Table~\ref{tab:e2e-cost} reports the end-to-end wall-clock cost on EgoLifeQA, including both memory construction and inference.
Compared with WorldMM, EM\textsuperscript{2}Mem increases the offline construction time from 88,708s to 99,022s, introducing an additional 10,314s construction overhead.
However, this extra offline cost is substantially smaller than the inference-time saving: EM\textsuperscript{2}Mem reduces the inference wall-clock time from 229,502s to 6,138s.
As a result, the overall end-to-end wall-clock time decreases from 318,210s to 105,160s, yielding a 3.03$\times$ speedup.

\begin{table}[H]
\centering
\footnotesize
\setlength{\tabcolsep}{3.2pt}
\renewcommand{\arraystretch}{0.96}
\begin{tabular*}{\columnwidth}{@{\extracolsep{\fill}}lrrr@{}}
\toprule
\textbf{Metric} & \textbf{WorldMM} & \textbf{EM\textsuperscript{2}Mem} & \textbf{$\Delta$ / Gain} \\
\midrule
\multicolumn{4}{@{}l}{\textit{Wall-clock time}} \\
Memory-Cons. (s) & 88,708 & 99,022 & +10,314 \\
Inference (s) & 229,502 & 6,138 & -223,364 \\
End-to-End (s) & 318,210 & \textbf{105,160} & \textbf{3.03$\times$} \\
\midrule
\multicolumn{4}{@{}l}{\textit{Token cost}} \\
Memory-Cons & 60.55M & 89.23M & +28.68M \\
Inference & 42.03M & 15.27M & -26.76M \\
Total & 102.58M & 104.50M & +1.92M \\
\bottomrule
\end{tabular*}
\caption{
End-to-end cost on EgoLifeQA. ``Memory-Cons.'' denotes offline memory construction, and ``Inference'' denotes QA inference.
EM\textsuperscript{2}Mem incurs higher construction cost but substantially reduces inference cost, yielding a 3.03$\times$ end-to-end wall-clock speedup.
The token-count break-even occurs after approximately 536 queries.
}
\label{tab:e2e-cost}
\end{table}

\paragraph{When does offline memory construction pay off.}
Let $\Delta T_{\mathrm{build}}$ denote the additional construction time of EM\textsuperscript{2}Mem over WorldMM, and let $\Delta T_{\mathrm{infer}}$ denote the per-query inference-time saving.
The number of queries required to amortize the additional construction cost is
\[
N_{\mathrm{break-even}}
=
\frac{\Delta T_{\mathrm{build}}}{\Delta T_{\mathrm{infer}}}.
\]
Using the measured average latency per query, EM\textsuperscript{2}Mem saves 360.79s per query, leading to a break-even point of approximately 29 queries.
Using the practical evaluation throughput measured on EgoLifeQA, the break-even point is approximately 24 queries.
This indicates that the proposed construction-time organization is beneficial when a processed long-video memory is reused for a moderate number of questions, such as benchmark evaluation, personal lifelog QA, surveillance-free archival search, or enterprise video repositories.
In contrast, for one-off queries over videos that will never be revisited, the additional construction cost may not be fully amortized.

\paragraph{Token-cost amortization.}
We also analyze the same trade-off in terms of token usage.
As shown in Table~\ref{tab:e2e-cost}, EM\textsuperscript{2}Mem uses 28.68M more tokens than WorldMM during memory construction, but saves 26.76M tokens during inference on the 500-query EgoLifeQA evaluation set.
Under an unweighted token-count calculation, this corresponds to a break-even point of approximately 536 queries.
Therefore, EM\textsuperscript{2}Mem is immediately advantageous in wall-clock efficiency, while its token-cost advantage becomes clearer when the constructed memory is reused across a larger number of queries.
Moreover, since EM\textsuperscript{2}Mem reduces output tokens more aggressively than input tokens during inference, the monetary break-even point can be lower when completion tokens are more expensive.

\subsection{Detailed Ablation Study}
\label{app:ablation}
\paragraph{Setup.}
Table~\ref{tab:category_ablation} reports category-wise ablation results on EgoLifeQA. 
We compare the full EM\textsuperscript{2}Mem with four variants: \textit{-TCV} removes temporal context views and only uses event-level evidence without coarser temporal aggregation; \textit{-SM} removes semantic memory, so long-term facts and habits are not used as supporting evidence; \textit{-EG} removes episodic graph evidence and disables graph-based event expansion; and \textit{30s} restricts retrieval to 30-second event records only.

\begin{table}[H]
\centering
\footnotesize
\setlength{\tabcolsep}{3.0pt}
\renewcommand{\arraystretch}{0.95}
\begin{tabular*}{\columnwidth}{@{\extracolsep{\fill}}lccccc@{}}
\toprule
\textbf{Type} 
& \textbf{Full} 
& \textbf{w/o TCV} 
& \textbf{w/o SM} 
& \textbf{w/o EG} 
& \textbf{30s-only} \\
\midrule
EntityLog 
& 60.8 & 57.6 & 56.8 & 56.8 & \textbf{61.6} \\
EventRecall 
& 61.1 & 60.3 & 59.5 & 58.7 & \textbf{65.1} \\
HabitInsight 
& \textbf{63.9} & 60.7 & 62.3 & 60.7 & 60.7 \\
RelationMap 
& \textbf{72.8} & 60.8 & 65.6 & 68.8 & 64.8 \\
TaskMaster 
& \textbf{74.6} & 65.1 & 65.1 & 63.5 & 68.3 \\
Overall 
& \textbf{66.0} & 60.4 & 61.4 & 61.6 & 64.0 \\
\bottomrule
\end{tabular*}
\caption{
Category-wise ablation on EgoLifeQA.
Full denotes EM\textsuperscript{2}Mem.
w/o TCV, w/o SM, and w/o EG remove temporal context views, semantic memory, and episodic graph, respectively.
30s uses only 30-second event records.
Accuracy is reported in \%.
}
\label{tab:category_ablation}
\end{table}

\paragraph{Analysis.}
The full model achieves the best overall accuracy, outperforming all ablated variants by 2.0--5.6 points. 
The gains are most evident on HabitInsight, RelationMap, and TaskMaster, where questions require recurring patterns, interpersonal relations, or task-level continuity beyond a single short event. 
Removing temporal context views causes the largest overall drop, showing the importance of coarse context for connecting sparse evidence across long videos. 
Removing semantic memory and episodic graph also degrades performance, indicating that long-term semantic support and event-level relational links are both useful for evidence-driven reasoning. 
The 30s-only variant performs best on EntityLog and EventRecall, suggesting that fine-grained event records are sufficient for local object-centric or direct recall questions, but its weaker results on long-range categories confirm the need for temporal context, episodic graph structure, and semantic memory.

\subsection{Analysis of Construction-Time Alignment}
\label{app:alignment_control}

WorldMM~\cite{WorldMM}, our strongest and most related memory-based baseline, also employs multi-scale, semantic, visual, and graph-based memories. 
To isolate the contribution of construction-time evidence alignment from these shared components, we compare two controlled variants on all 500 EgoLifeQA questions using the same 6,057 underlying 30-second records. 
Both use identical caption, transcript, visual, and structured evidence, retrieval and answer models, prompts, causal filtering, and temporal budgets, while disabling multi-scale context, graph expansion, LLM selection, and query planning. 
\textit{Four-channel RRF} indexes the four evidence channels independently and combines their rankings with reciprocal-rank fusion, whereas \textit{Aligned-30s} binds the same fields into one provenance-linked record per temporal anchor before indexing.

\begin{table}[H]
\centering
\footnotesize
\setlength{\tabcolsep}{2pt}
\renewcommand{\arraystretch}{0.95}

\begin{tabular}{@{}lrrrrrr@{}}
\toprule
\textbf{Memory}
& \multirow{2}{*}{\textbf{$R@1$}}
& \multirow{2}{*}{\textbf{$R@5$}}
& \multirow{2}{*}{\textbf{$C@150$}}
& \multirow{2}{*}{\textbf{$C@300$}}
& \multirow{2}{*}{\textbf{$C@600$}}
& \multirow{2}{*}{\textbf{QA}} \\
\textbf{organization}
& & & & & & \\
\midrule
4-channel RRF
& 8.9 & 24.2 & 24.3 & 31.2 & 41.1 & 60.0 \\
Aligned-30s
& \textbf{14.0}
& \textbf{27.6}
& \textbf{27.9}
& \textbf{36.7}
& \textbf{44.6}
& \textbf{63.0} \\
Gain
& +5.1 & +3.4 & +3.6 & +5.4 & +3.5 & +3.0 \\
\bottomrule
\end{tabular}

\caption{Controlled comparison of independent four-channel retrieval and construction-time aligned retrieval on all 500 EgoLifeQA questions. $R@k$ denotes top-$k$ target-time recall, $C@B$ denotes macro target-time coverage under a $B$-second retrieved-video budget, and QA denotes final-answer accuracy (\%).}
\label{tab:alignment_control}
\end{table}
\begin{table*}[t]
\centering
\small
\setlength{\tabcolsep}{5.0pt}
\renewcommand{\arraystretch}{0.95}
\begin{tabular*}{\textwidth}{@{\extracolsep{\fill}}lccccccc@{}}
\toprule
\textbf{Question Type} 
& \textbf{\#Questions} 
& \textbf{Cap.-Late} 
& \textbf{Cap.-Early} 
& \textbf{Vis.-Late} 
& \textbf{Vis.-Early} 
& \textbf{SER-Late} 
& \textbf{EM\textsuperscript{2}Mem} \\
\midrule
EntityLog    
& 54  
& 59.3 
& 63.0 
& 57.4 
& 64.8 
& \textbf{68.5} 
& 64.8 \\
EventRecall  
& 45  
& 68.9 
& 68.9 
& 66.7 
& 68.9 
& 68.9 
& 68.9 \\
HabitInsight 
& 40  
& 65.0 
& 65.0 
& \textbf{72.5} 
& 67.5 
& 62.5 
& 65.0 \\
RelationMap  
& 71  
& 66.2 
& 64.8 
& 70.4 
& 73.2 
& 69.0 
& \textbf{74.6} \\
TaskMaster   
& 40  
& 67.5 
& 67.5 
& 67.5 
& 67.5 
& 70.0 
& \textbf{77.5} \\
\midrule
Overall      
& 250 
& 65.2 
& 65.6 
& 66.8 
& 68.0 
& 68.0 
& \textbf{71.2} \\
\bottomrule
\end{tabular*}
\caption{
Analysis of evidence interface and alignment timing on EgoLifeQA.
Cap., Vis., and SER denote flat caption evidence, raw visual evidence, and structured event records, respectively.
Late denotes retrieval-time alignment, while Early denotes construction-time alignment.
EM\textsuperscript{2}Mem denotes our event-centric unified multimodal memory framework.
All numbers denote accuracy (\%).
}
\label{tab:evidence_unification}
\end{table*}

As shown in Table~\ref{tab:alignment_control}, construction-time alignment consistently improves evidence localization and coverage. 
Aligned-30s improves $R@1$ and $R@5$ by 5.1 and 3.4 points, respectively, and increases coverage across all temporal budgets. 
At the primary 300-second budget, coverage improves by 5.4 points (95\% CI: [+2.6, +8.3]; McNemar exact $p=0.0003$). 
Final QA accuracy also increases from 60.0\% to 63.0\%, although this difference is not statistically significant (95\% CI: [-0.8, +6.8]; $p=0.1505$). 
These results indicate that EM\textsuperscript{2}Mem's advantage is not solely attributable to auxiliary memory components shared with prior systems: binding heterogeneous evidence into a common retrieval unit before indexing improves evidence retrieval under matched conditions.

\subsection{Category-wise Evidence Unification Analysis}
\label{app:unification}
This section extends Section~\ref{sec:analysis} with category-wise results on EgoLifeQA. 
Table~\ref{tab:evidence_unification} shows that EM\textsuperscript{2}Mem achieves the best overall accuracy, but the benefit is not uniform across question types. 
The largest gains appear on \textit{RelationMap} and \textit{TaskMaster}, where the model must connect people, actions, tasks, and temporally distant evidence across multiple events. 
This suggests that the main advantage of construction-time unification is not simply adding visual information, but turning heterogeneous cues into event-centered memory units that can be retrieved and composed as coherent evidence chains. 
In contrast, \textit{EventRecall} is relatively insensitive to the choice of evidence interface, indicating that direct recall questions can often be answered from localized evidence without requiring strong cross-event integration.

The results also reveal the boundary of structured memory. 
Structured event records perform strongly on \textit{EntityLog}, suggesting that typed fields such as objects, actions, and scenes provide an effective interface for local entity-centric reasoning. 
However, raw visual evidence performs best on \textit{HabitInsight}, implying that some recurring perceptual cues may be weakened when visual information is compressed into textual or structured fields. 
This observation supports our final design: EM\textsuperscript{2}Mem uses structured event records as the primary retrieval interface for efficient and traceable reasoning, while retaining raw keyframes only for targeted visual verification after relevant event-centric multimodal memory selection. 
Overall, the category-wise trends refine the main conclusion in Section~\ref{sec:analysis}: effective multimodal memory depends on both \textit{how} evidence is represented---as structured event-centered memory---and \textit{when} it is unified---before retrieval rather than during inference.

\begin{figure}[!htbp]
    \centering
    \includegraphics[width=0.5\textwidth]{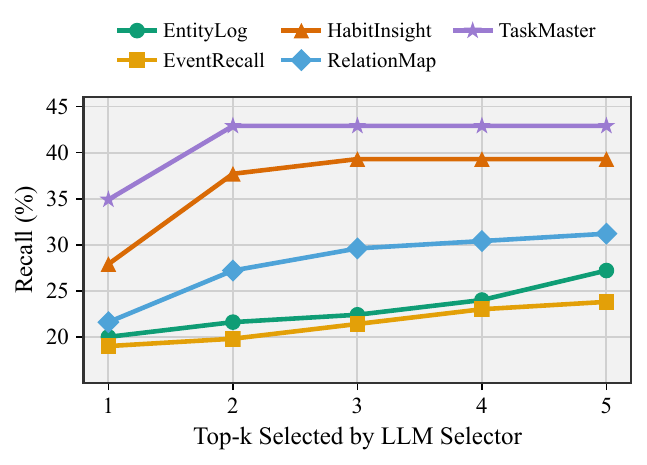}
    \caption{Event-level recall under different selector budgets.}
    \label{fig:selector-topk-recall}
\end{figure}

\begin{figure*}[t]
    \centering
    \includegraphics[width=\textwidth]{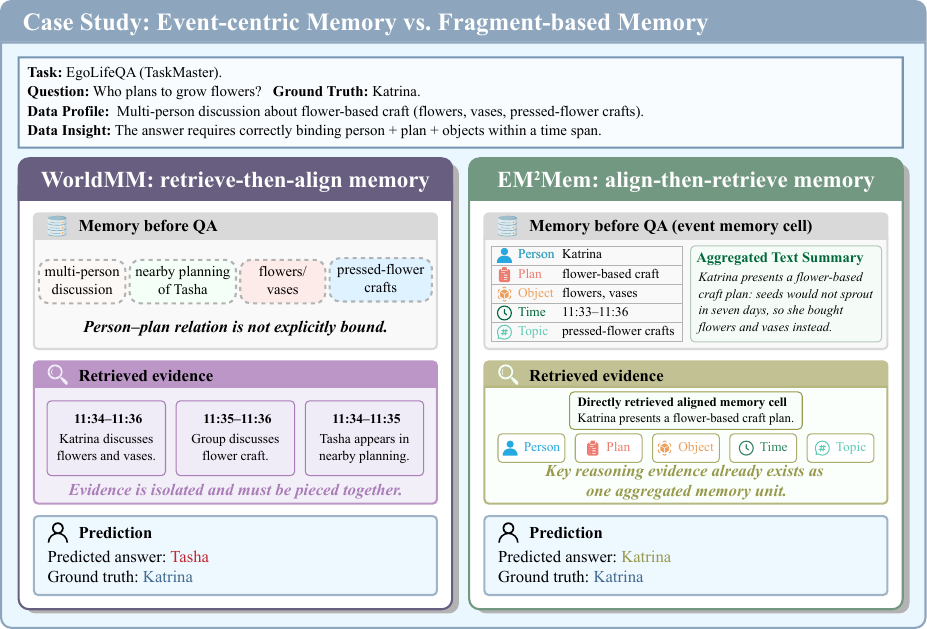}
    \caption{Case study of TaskMaster in EgoLifeQA.}
    \label{fig:case_study}
\end{figure*}

\subsection{Selector budget}
\label{app:selector-budget}
We evaluate whether EM\textsuperscript{2}Mem retrieves the annotated temporal evidence more effectively than WorldMM.
We use strict 30-second event-level recall, where a prediction is counted as correct only when the retrieved event anchor matches the annotated evidence segment.
Table~\ref{tab:compact_recall} and Figure~\ref{fig:selector-topk-recall} compares WorldMM's cumulative recall after 1 and 5 iterative retrieval rounds with EM\textsuperscript{2}Mem's single-pass Top-\(k\) event-anchor retrieval.
EM\textsuperscript{2}Mem achieves 30.8\% Top-5 recall, outperforming WorldMM 5R by 7.0 points.
EM\textsuperscript{2}Mem Top-1 recall also reaches 23.0\%, close to WorldMM 5R at 23.8\%.
The gains are larger on HabitInsight and TaskMaster, suggesting that event-indexed memory is helpful when evidence depends on long-term patterns or task states.
EventRecall is the only category where EM\textsuperscript{2}Mem is slightly lower, indicating that localized event-recall questions may still benefit from iterative caption-level retrieval.
Overall, these results suggest that EM\textsuperscript{2}Mem improves evidence localization for structured and long-range reasoning while preserving competitive localized recall.

\subsection{Anchor Granularity and Boundary Sensitivity}
\label{app:anchor_sensitivity}
The base anchor in EM\textsuperscript{2}Mem is a fixed temporal indexing address rather than a learned semantic boundary. 
We evaluate its granularity on all 500 EgoLifeQA questions by comparing fixed 30-, 60-, and 120-second windows with two variable-length proxies that merge consecutive 30-second records while the generated scene label or speaker set remains unchanged. 
All variants use the same evidence and retrieval settings. 
Since longer windows cover more video per retrieved item, we report macro target-time coverage under equal retrieved-video budgets of 150, 300, and 600 seconds.

\begin{table}[H]
\centering
\small
\setlength{\tabcolsep}{6pt}
\renewcommand{\arraystretch}{0.95}
\begin{tabular}{lccc}
\toprule
\textbf{Anchor construction}
& \textbf{C@150}
& \textbf{C@300}
& \textbf{C@600} \\
\midrule
Fixed 30s
& \textbf{27.9}
& \textbf{36.7}
& \textbf{44.6} \\
Fixed 60s
& 20.2
& 30.3
& 39.4 \\
Fixed 120s
& 16.5
& 23.9
& 34.4 \\
Scene-boundary proxy
& 17.6
& 26.0
& 36.2 \\
Speaker-boundary proxy
& 26.8
& 33.3
& 42.3 \\
\bottomrule
\end{tabular}
\caption{Anchor sensitivity on all 500 EgoLifeQA questions.
$C@B$ denotes macro target-time coverage (\%) under a
$B$-second retrieved-video budget.}
\label{tab:anchor_sensitivity}
\end{table}

As shown in Table~\ref{tab:anchor_sensitivity}, fixed 30-second anchors achieve the highest coverage at all budgets; at 300 seconds, they outperform the alternatives by 3.4--12.8 points.
We also tested identity-based adaptive boundaries using YOLOv8 with BoT-SORT on one full day of EgoLife video.
Despite only six protagonists, tracking produced 6,130 person IDs, with frequent fragmentation under egocentric motion, occlusion, and re-entry.
These results suggest that 30-second anchors offer a robust localization--budget trade-off on EgoLifeQA, without implying a universal semantic event boundary.

\begin{table*}[t]
\centering
\footnotesize
\setlength{\tabcolsep}{5pt}
\renewcommand{\arraystretch}{1.08}
\begin{tabularx}{\textwidth}{@{}>{\raggedright\arraybackslash}p{0.15\textwidth}>{\raggedright\arraybackslash}X>{\raggedright\arraybackslash}p{0.24\textwidth}@{}}
\toprule
\textbf{Stage or object} & \textbf{Actual stored or retrieved evidence} & \textbf{Role in the trace} \\
\midrule
Local record
& At \texttt{DAY1\_11333000\_11340000}, Katrina says that she bought flowers and vases, that the flowers will bloom in 1--2 days, and that this fits the 7-day cycle. A linked keyframe independently shows pink flowers in a vase.
& Binds the speaker, intended activity, time span, and visual observation to one source interval. \\

Episodic graph
& The node \texttt{event::DAY1\_11333000\_11340000} \texttt{involves} Katrina, is \texttt{about} buying flowers and vases, and \texttt{occurs\_in} the dining area. Extracted triplets include (Katrina, \texttt{say\_about}, bought flowers and vases) and (Katrina, \texttt{say\_about}, fits 7-day cycle).
& Preserves the person--plan association. Every relation retains its event ID, document ID, and 30s scale for source tracing. \\

Semantic distractors
& Retrieved facts also state that Tasha says seed paper can be planted (support count 5) and that sunflower seeds are easier to sprout (support count 1).
& These facts overlap with \textit{grow flowers}, but do not identify the owner of the queried plan; they enter only through their provenance-linked anchors. \\

Ranking and selection
& With a causal cutoff at 11:57:36, the selector retains five anchors from the 12-anchor pool: 11:35:30 (1.0000), 11:47:30 (0.9288), 11:47:01 (0.8839), 11:48:00 (0.4272), and 11:34:30 (0.3954).
& The scores are produced by coarse event ranking; bounded selection compares the complete event packets without searching the full memory. \\

Grounded readout
& The 3-minute view around 11:35 contains Katrina's 7-day cultivation plan, while the 11:48 event adds her statement, ``We can plant it too.'' Later Alice events concern bulbs purchased as prizes rather than her own growing plan.
& The assembled evidence resolves plan ownership and yields the correct answer, \textbf{D: Katrina}. \\
\bottomrule
\end{tabularx}
\caption{Concrete memory objects and query-time trace for the question ``Who plans to grow flowers?'' Scores in the ranking row are normalized coarse anchor scores.}
\label{tab:case_study_trace}
\end{table*}

\subsection{Case Study}
\label{app:case-study}
Figure ~\ref{fig:case_study} presents a qualitative case study on an EgoLifeQA TaskMaster question: ``Who plans to grow flowers?'' 
The answer requires identifying plan ownership in a multi-person discussion where flowers, vases, craft ideas, and nearby mentions of Tasha appear together. 
WorldMM retrieves several flower-related snippets, but they remain loosely connected textual fragments without an explicit person--plan binding, leading to the wrong prediction Tasha. 
In contrast, EM\textsuperscript{2}Mem aligns multimodal evidence during memory construction and stores it as an event-centric memory cell, where Katrina is explicitly linked to the flower-based plan, associated objects, temporal context, and supporting visual evidence.

Table~\ref{tab:case_study_trace} further traces how this evidence propagates through the pipeline. 
The local record and episodic graph preserve Katrina-specific person--plan relations with event provenance, while retrieved Tasha-related semantic facts provide plausible but non-decisive distractors. 
After causal ranking and bounded selection, the retained Katrina events contain both the 7-day cultivation context and the explicit statement ``We can plant it too,'' whereas Alice-related evidence concerns bulbs purchased as prizes. 
The grounded readout therefore resolves the ambiguous flower mentions to Katrina and yields the correct answer. 
This example illustrates how event-centric alignment, provenance-linked graph structure, and bounded retrieval jointly turn heterogeneous evidence into a query-ready evidence unit.

\section{Extended Related Work}
\label{app:extended_related_work}
\paragraph{Long Video Understanding.}
Recent video-language models have achieved substantial progress in processing extended visual sequences. 
Early video-language pretraining methods learn transferable visual-semantic representations through joint video-text
modeling, multimodal alignment, and masked spatiotemporal modeling \cite{VideoBert, ALPRO, VideoMAE, CLIP-ViP, InternVideo2}.
Long-form and temporal-aware pretraining further improve video-language modeling over extended temporal contexts \cite{LF-VILA, HiTeA, EMCL, mPLUG, Champagne}. 
Building on these foundations, video large language models connect visual encoders with LLMs and improve instruction-following through multimodal instruction tuning and unified image-video representations \cite{Video-LLaMA, Video-ChatGPT, Video-LLaVA, Chat-UniVi, LLaVA-OneVision}. 
Recent systems further strengthen caption quality, audio-spatiotemporal modeling, and image-video encoder integration \cite{ShareGPT4Video, VideoLLaMA_2, VideoGPT+, VideoLLaMA_3, Tarsier}.
More recent work extends video LLMs to longer contexts by
compressing visual tokens, maintaining sparse memory, or adapting visual inputs to long-context LLMs \cite{MovieChat, LLaMA-VID, LongVA, LongVU, VideoChat-Flash}. 
Other studies improve temporal localization and reasoning
through time-aware modeling, temporal grounding, or test-time reasoning \cite{TimeChat, Time-r1, Video-RTS, Video-Thinker}. 
Nevertheless, as video duration increases from minutes to hours or days, task-relevant evidence often becomes sparse, temporally distant, and difficult to localize, motivating explicit mechanisms for organizing and retrieving video-specific information.

\paragraph{Multimodal Memory for Large Language Models.}
To enable scalable reasoning over long videos, recent studies have explored explicit memory and retrieval mechanisms. Retrieval-augmented methods build external memories from captions, transcripts, OCR, keyframes, clip embeddings,
or graph-structured indices, and retrieve query-relevant evidence before answer generation \cite{LightRAG, HippoRAG, Video-RAG__Visually-aligned, VideoRAG__Retrieval-Augmented, LangRepo}. 
Other methods organize videos into document-like, hierarchical, or adaptive structures for efficient long-video
reasoning \cite{DrVideo, VideoTree, OneClip-RAG, StructMem, HAVEN}. 
In egocentric long-video understanding, EgoRAG constructs multi-level caption and summary memories for coarse-to-fine retrieval \cite{EgoLife, MobileMem, deng2026mobilemem}, while HippoMM segments audiovisual
streams into episodes and consolidates them into semantic summaries for hierarchical recall \cite{HippoMM}. 
Beyond fixed retrieval pipelines, agentic memory systems incorporate memory access into multi-step reasoning
for temporal localization, multimodal inspection, and iterative evidence gathering \cite{VideoAgent, M3-Agent, Ego-R1, WorldMM, VideoARM, LycheeCluster}. 
Collectively, this line of work \cite{EventMemAgent, OASIS, MM-MeM, LightMem, TokenPilot, Dynamic_Long_Context_Reasoning, OneMind} motivates the development of unified memory systems capable of supporting fine-grained event grounding, flexible temporal abstraction, and long-term semantic reasoning \cite{MMSkills, DBLP:conf/acl/QiaoO0CYDTHC23, PersonaVLM, AutoDreamer, LightMem_Ego, Metis, lycheememoryv2}.

\FloatBarrier

\end{document}